\pdfoutput=1

\documentclass[11pt]{article}

\usepackage[final]{acl}

\usepackage{times}
\usepackage{latexsym}

\usepackage[T1]{fontenc}

\usepackage[utf8]{inputenc}

\usepackage{microtype}

\usepackage{inconsolata}

\usepackage{graphicx}

\usepackage{times}
\usepackage{latexsym}
\usepackage{amsmath}
\usepackage{amsfonts}
\usepackage[T1]{fontenc}
\usepackage[utf8]{inputenc}
\usepackage{microtype}
\usepackage{inconsolata}
\usepackage{graphicx}
\usepackage{xcolor}
\usepackage{tabularx}
\usepackage{colortbl}
\usepackage{ulem} %
\usepackage{subcaption}
\usepackage{booktabs}
\usepackage{siunitx}
\usepackage{multirow}
\usepackage{adjustbox, arydshln }

\newcommand\blfootnote[1]{%
  \begingroup
  \renewcommand\thefootnote{}\footnote{#1}%
  \addtocounter{footnote}{-1}%
  \endgroup
}

\definecolor{firstcolor}{RGB}{204, 212, 230}
\definecolor{secondcolor}{RGB}{241, 219, 206}
\definecolor{thirdcolor}{RGB}{210, 228, 212}
\definecolor{fourthcolor}{RGB}{250, 210, 210}
\definecolor{bordergray}{RGB}{150, 150, 150} %

\newcommand{\firstbox}[1]{\colorbox{firstcolor}{#1}}
\newcommand{\secondbox}[1]{\colorbox{secondcolor}{#1}}
\newcommand{\thirdbox}[1]{\colorbox{thirdcolor}{#1}}
\newcommand{\fourthbox}[1]{\colorbox{fourthcolor}{#1}}

\definecolor{darkblue}{rgb}{0, 0, 0.5}
\hypersetup{colorlinks=true, citecolor=darkblue, linkcolor=darkblue, urlcolor=darkblue}

\usepackage[compact]{titlesec} %
\title{Evil twins are not that evil:\\Qualitative insights into machine-generated prompts}

\author{
 \textbf{Nathanaël Carraz Rakotonirina\textsuperscript{* 1}},
 \textbf{Corentin Kervadec\textsuperscript{* 1}},
 \textbf{Francesca Franzon\textsuperscript{1}}, \\
 \textbf{Marco Baroni\textsuperscript{1,2}}
\\
 \textsuperscript{1}Universitat Pompeu Fabra, Barcelona \\
 \textsuperscript{2}ICREA, Barcelona
}

\begin{document}
\maketitle

\begin{abstract}
 It has been widely observed that language models (\textit{LMs}) respond in predictable ways to algorithmically generated prompts that are seemingly unintelligible. This is both a sign that we lack a full understanding of how LMs work, and a practical challenge, because opaqueness can be exploited for harmful uses of LMs, such as jailbreaking. We present the first thorough analysis of opaque machine-generated prompts, or \textit{autoprompts}, pertaining to 6 LMs of different sizes and families. We find that machine-generated prompts are characterized by a last token that is often intelligible and strongly affects the generation. A small but consistent proportion of the previous tokens are prunable, probably appearing in the prompt as a by-product of the fact that the optimization process fixes the number of tokens. The remaining tokens fall into two categories: filler tokens, which can be replaced with semantically unrelated substitutes, and keywords, that tend to have at least a loose semantic relation with the generation, although they do not engage in well-formed syntactic relations with it. Additionally, human experts can reliably identify the most influential tokens in an autoprompt \textit{a posteriori}, suggesting these prompts are not entirely opaque. Finally, some of the ablations we applied to autoprompts yield similar effects in natural language inputs, suggesting that autoprompts emerge naturally from the way LMs process linguistic inputs in general. 
\end{abstract}

\section{Introduction}
\label{sec:introduction}

\blfootnote{* Equal contribution. Correspondence:  \href{mailto:nathanael.rakotonirina@upf.edu}{nathanael.rakotonirina@upf.edu}.}

\begin{figure*}[th]
    \centering
    \includegraphics[width=0.79\linewidth]{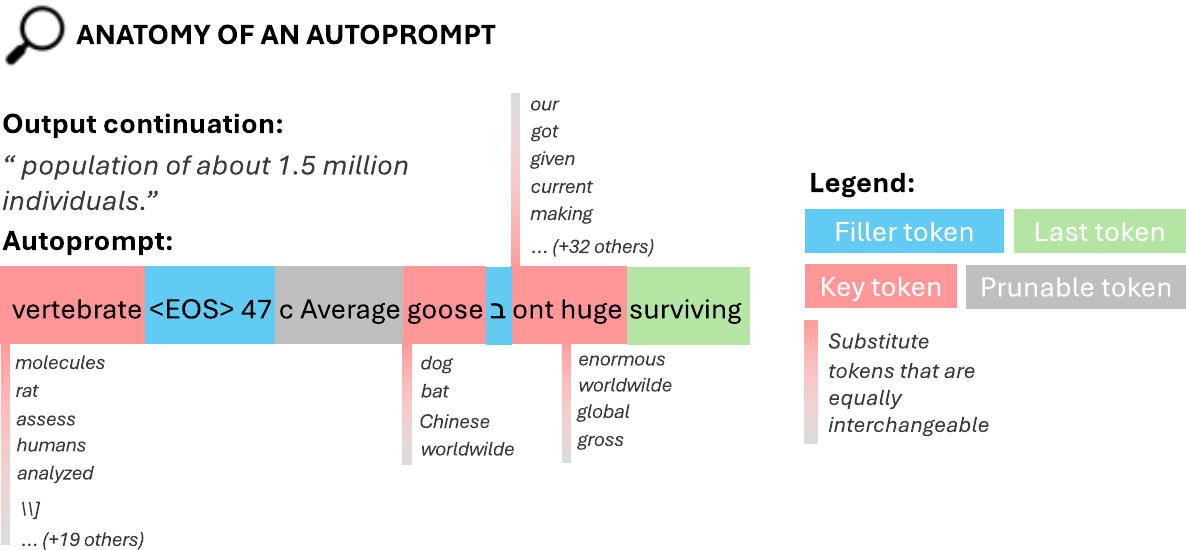}
    \caption{We analyze opaque machine-generated prompts (autoprompts) and identify four key components: (1) the \textit{last token}, highly influential and difficult to modify; (2) \textit{prunable} tokens, being ignored by the LM; (3) \textit{key tokens}, which carry loosely related semantic information, essential for generation, and that can be replaced with semantically similar tokens; and (4) \textit{fillers}, which can be substituted with a large amount of unrelated tokens but, unlike prunable tokens, cannot be deleted.}
    \label{fig:teaser}
\end{figure*}

An intriguing property of language models (\textit{LMs}) is that they respond in predictable ways to machine-generated prompts (henceforth, \textit{autoprompts})\footnote{The term \textit{autoprompt} was coined by \citet{Shin:etal:2020} to refer to the prompts generated by their algorithm. We repurpose the term here to refer to machine-generated prompts in general.}
that are unintelligible to humans. \citet{Shin:etal:2020} first showed that autoprompts can outperform human-crafted prompts on various tasks. More worryingly, \citet{Wallace:etal:2019} and others have shown that they can be used in adversarial attacks making models, including latest-generation aligned LMs, behave in undesirable ways  \citep[e.g.,][]{Zou:etal:2023,Geiping:etal:2024}. We present here the first thorough qualitative analysis of autoprompts. We discover that, despite the superficial impression of opacity they convey, they can to a significant extent be explained in terms of a few general observations (illustrated in Figure~\ref{fig:teaser}):
(1) in autoregressive models, the last token of a prompt has a disproportionate role in generating the continuation, and this last token is both very important and often transparent in autoprompts; 
(2) several tokens contributing to the opaqueness of autoprompts are simply ignored by the model;
(3) the non-final elements that are actually influencing generation might do so in two ways: interchangeable tokens acting as fillers, and more semantically coherent keywords tokens. 
As we will see, these factors are also at play when LMs are fed natural-language sequences, suggesting that they are core properties of how LMs process linguistic strings.

From a theoretical point of view, our study offers new insights into LM language processing in general. From a practical point of view, it highlights which aspects of LMs we should pay attention to, if we want to make them more robust to harmful autoprompts (or, conversely, to develop more efficient benign autoprompt generation techniques). We present the first thorough analysis of opaque machine-generated prompts, or autoprompts, pertaining to 6 LMs of different sizes and families, focusing on minimal pairs of natural language prompts and their “evil twins”, i.e., opaque autoprompts that lead to the same continuation.

\section{Related work}
\label{sec:related}

Starting with the seminal work of \citet{Wallace:etal:2019} and \citet{Shin:etal:2020}, many studies have revealed that, using various discrete gradient-following techniques, it is possible to automatically discover prompts that, while unintelligible, let LMs generate a desired target output \citep[e.g.,][]{Deng:etal:2022,Wen:etal:2023}. Moreover, such prompts are at least to some degree transferable, in the sense that they can be induced using a LM, but then successfully used to prompt a different one, including much larger models \citep{Rakotonirina:etal:2023,Zou:etal:2023}. Initially, the interest was mainly in whether algorithmically-generated autoprompts could be used as alternatives to manually crafted prompts in knowledge-extraction tasks or other applications, but with recent progress in LM's ability to respond to natural language prompts, this goal has become somewhat obsolete. Autoprompts are however still an important concern because they can be used for adversarial purposes, for example to bypass safety filters to generate offensive or dangerous information \cite[e.g.,][]{Zou:etal:2023,Geiping:etal:2024}. Even more importantly, the fact that several modern LMs are more likely to provide information about the star formation process when prompted with \texttt{``Produ bundcules cation of\` stars efect''} than when prompted with the question ``What leads to the creation of new stars?'' suggests that we still do not understand something fundamental about how LMs process language \citep{Melamed:etal:2024}.

There is relatively little work attempting to characterize the nature of autoprompts. \citet{Geiping:etal:2024} present a set of intriguing qualitative observations about how autoprompts support various types of attacks (e.g., by including instruction fragments in different languages), as well as an analysis of tokens commonly appearing in autoprompts. \citet{Ishibashi:etal:2023} find that autoprompts are less robust to token re-arrangement than natural prompts, whereas \citet{Rakotonirina:etal:2023} report that the autoprompts that best transfer across models contain a larger proportion of English words and, surprisingly, are \textit{less} order-sensitive than  autoprompts that do not transfer. \citet{Kervadec:etal:2023} analyze the activation paths of autoprompts and comparable natural sequences across the layers of a LM, finding that often they follow distinct pathways. \citet{Melamed:etal:2024} study, like us, what they call ``evil twins'', namely autoprompts that produce continuations comparable to those of a reference natural sequence. They compare the relative robustness to token shuffling of autoprompts and natural prompts, finding that, depending on the model family, autoprompts might be more, less or comparably robust to shuffling. They also run a substitution experiment similar to the one we will describe below (but replacing tokens with a single, fixed, [UNK] token). They find that this ablation strongly affects the autoprompts: we find a more nuanced picture, by considering a large range of possible replacements.

\section{Experimental setup}
\label{sec:methods}

\paragraph{Models} We use decoder-only LMs from the Pythia \citep{Biderman:etal:2023} and OLMo \citep{Groeneveld:etal:2024} families, as these are fully open-source models whose training data are publicly available. Specifically, in the text we discuss the results we obtained with Pythia-6.9B, and we replicate the main experiments with Pythia-1.4B, Pythia-12B, OLMo-1B, OLMo-7B, and OLMo-7B-Instruct in App. \ref{app:other-models}, reporting similar results.

\arrayrulecolor{gray!50}
\begin{table*}[tb]
  \centering
  \small
  \resizebox{\textwidth}{!}{%
  \begin{tabular}{ll}
    \specialrule{1pt}{0pt}{0pt}
        \rowcolor{gray!20} \textbf{Autoprompts} & \textbf{Generated Continuation}   \\
    \specialrule{1pt}{0pt}{2pt}
    pulls\textbf{proper} Ryan\textbf{SP 184} critics \textbf{Mat?} embry " &  The film is a mess, but it's a mess that's worth seeing. \\
    autoimmune,"antibodies?\textbf{*<EOT>}arthyhatic:\_ they & attack the body's own tissues. \\
    \#\#\#\textbf{iotics} parental \textbf{=} depressive teen ? lossJulies & parents are going through a divorce. \\
    \specialrule{1pt}{0pt}{0pt}
        \rowcolor{gray!20} \textbf{Original Prompts} & \textbf{Generated Continuation} \\
    \specialrule{1pt}{0pt}{2pt}
    \ldots{} \textbf{Aviation} Regiment (\textbf{based in Giebel}stadt, & Germany) landed at the airport. \\
    \ldots{} \textbf{Robert Humanick of Slant Magazine} wrote\textbf{,} "& The film is a mess, but it's a mess that's worth seeing. \\ 
    \ldots{} appealed to \textbf{the Government for }additional funding, a third & of which would come from the Treasury. \\
    \ldots{} \textbf{an} autoimmune reaction causing \textbf{the body's immune} cells to & attack the body's own tissues. \\
    \ldots{} \textbf{)} — try to lead her through life as her & parents are going through a divorce. \\
    \bottomrule
  \end{tabular}
  }
  \caption{Randomly selected examples of autoprompts and original prompts for Pythia-6.9B, with prunable tokens in bold. For original prompts, only the last 10 tokens are shown. '?' = difficult-to-render characters.}
  \label{tab:merged-pruning-examples}
\end{table*}
\arrayrulecolor{black}

\arrayrulecolor{gray!50}
\begin{table*}[tb]
  \centering
  \small
  \begin{adjustbox}{width=1\textwidth}
  \begin{tabular}{llllll|rrrrr}
    \specialrule{1pt}{0pt}{0pt}
        \rowcolor{gray!20} & \multicolumn{5}{c|}{\textbf{Autoprompt}} & \multicolumn{5}{c}{\textbf{Original Prompt}} \\
    \specialrule{1pt}{0pt}{2pt}
    \multirow{2}{3em}{\textbf{Kept}}& \firstbox{British} & \secondbox{-} & \secondbox{'} & \secondbox{v} & \firstbox{led}  & \thirdbox{is} & \secondbox{)} & \thirdbox{after} & \secondbox{"} & \secondbox{),} \\
    & \firstbox{King} & \firstbox{West} & \firstbox{remained} & \firstbox{inaugural} & \firstbox{five} & \thirdbox{be} & \thirdbox{which} & \secondbox{(} & \thirdbox{she} & \firstbox{film} \\
    \midrule
    \multirow{2}{3em}{\textbf{Pruned}} & \secondbox{(} & \thirdbox{was} & \secondbox{.} & \thirdbox{for} & \thirdbox{The} & \thirdbox{the} & \secondbox{,} & \thirdbox{of} & \thirdbox{a} & \thirdbox{In} \\
    & \thirdbox{on} & \thirdbox{In} & \secondbox{âĢĶ} & \thirdbox{be} & \thirdbox{not} & \secondbox{.} & \secondbox{:} & \thirdbox{for} & \thirdbox{been} & \secondbox{a} \\
    \bottomrule
  \end{tabular}
  \end{adjustbox}
  \caption{Top-10 kept or pruned tokens for Pythia-6.9B, ranked by local mutual information for autoprompt and original prompts (for each cell, top-left has the highest value and bottom-right the lowest). Tokens are printed as follows: \firstbox{content word}, \secondbox{artifact and punctuation}, \thirdbox{function word}}
  \label{tab:lmi-merged}
\end{table*}
\arrayrulecolor{black}

\paragraph{Data collection} We sample 25k random English sequences from the WikiText-103 corpus \citep{Merity:etal:2016}, such that they contain between 35 and 80 (orthographic) tokens, and they are not interrupted by sentence boundary markers. We refer to these corpus-extracted sequences as \textit{original prompts}. We also record the original continuation of these sequences in the corpus. We let moreover the LM generate a continuation of each prompt using greedy decoding. The generation process stops after a maximum of 25 tokens or when end-of-sentence punctuation is encountered. We filter out sequences whose generated continuation is less than 4 tokens long. As we are interested in genuine model generation, as opposed to cases where the model is simply producing a memorized corpus sequence, we compute the BLEU score \citep{Papineni:etal:2002}\footnote{We use a modified version of BLEU that does not penalize short sequences. Scores are computed for up to 4-grams using uniform weights and add-$\epsilon$ smoothing.} between the model continuation and the original continuation, removing sequences with BLEU greater than 0.1.\footnote{\citet{Schwarzschild:etal:2024} find that sometimes autoprompts act as ``keys'' to retrieve memorized materials. This is an intriguing property we don't further explore here, as we're interested in their more general ability to generate natural-language sequences.} After filtering, we are left with a total of 5k sequences, which we use to train autoprompts. This dataset allows us to generate more complex prompts, that target full sentences instead of unique tokens. 

\paragraph{Prompt optimization} For each target continuation, we want to find a fixed-length autoprompt that makes the model produce that continuation. To achieve that, we maximize the probability of the target continuation given the prompt. More formally, if we denote the target sequence by $(t_1,...,t_m) \in \mathcal{V}^m$, where $\mathcal{V}$ is the vocabulary, and the $n$-length autoprompt by $(p_1,...,p_n) \in \mathcal{V}^n$  (in our case, $n=10$), the optimization problem can be formulated as follows:\footnote{We empirically observed that using more than 10 tokens only increases the number of useless tokens (cf. pruning experiment in Section~\ref{sec:pruning}) without introducing any distinctive features. On the contrary, using less than $10$ tokens was usually not enough to find the target continuation.}
\[\underset{(p_1,...,p_n) \in \mathcal{V}^n}{\text{minimize}} -\log \mathbb{P}_{LLM}(t_1,...,t_m|p_1,...,p_n)\]

We use a variant of Greedy Coordinate Gradient (GCG) \citep{Zou:etal:2023}, a widely used gradient-based algorithm that iteratively updates the prompt one token at a time \citep{Ebrahimi:etal:2018, Wallace:etal:2019, Shin:etal:2020}. During each iteration, we select the top 256 tokens with the largest negative gradients for every position, then we uniformly sample 256 candidates across all positions.  We then compute the loss of each candidate replacement, and select the one with the lowest loss. We run up to 50 iterations of this process.%
We discard cases in which, after these iterations, we have not found an autoprompt that produces the very same continuation.%

\paragraph{Data-set statistics} The final data-set we use for the Pythia-6.9B experiments reported in the main text consists of 208 triples of original prompt, autoprompt and continuation.\footnote{We study relatively few autoprompts as it is very time-consuming to extract them for large model. Replicating the experiment with larger autoprompt sets using smaller models led to comparable results (App. \ref{app:other-models}).} The average original prompt length is of 39.3 tokens (s.d.~13.4); that of the continuations is of 8.4 tokens (s.d.~2.4).\footnote{Datasets and code are uploadedas supplementary materials, and will be made available upon publication.}

\section{Experiments}

\subsection{Pruning autoprompts}
\label{sec:pruning}

\paragraph{Methodology} We greedily prune the autoprompts in our data-set. Starting from the original sequence of \textit{n} tokens, we strip each token in turn, and pick the \textit{n-1}-length sequence that produces the same continuation as the original, if any (if there's more than one such sequence, we randomly pick one). We repeat the process starting from the shortened sequence, and stop where there is no shorter sequence generating the original continuation, or when we are down to a single-token prompt.

\paragraph{Roughly 20\% of the tokens are \textit{prunable}} It is possible to shorten the original autoprompt in a clear majority of the cases (73.2\%), with the average pruned autoprompt having lost 2.6 tokens of 10 (s.d.: 1.6). Table \ref{tab:merged-pruning-examples} (top section) shows randomly picked examples with the pruned tokens highlighted in bold. Autoprompt-discovery algorithms fix the number of tokens as a hyperparameter. It is thus reasonable that some tokens in the final autoprompt are just there to fill all the required slots, and can consequently be pruned. This view is supported by the following observation. We roughly classified the autoprompt tokens into \textit{language-like} and \textit{non-linguistic}, such as digits, punctuation, code-fragments and non-ascii characters.
We found that the proportion of non-linguistic tokens is decidedly higher among pruned tokens (46.6\%) than among kept tokens (24.1\%).\footnote{As a side note, we found that 28.4\% of the full-autoprompt tokens are non-linguistic.}
Table \ref{tab:lmi-merged} (left) further shows tokens that are most typically kept or removed by the pruning algorithm according to the local mutual information statistics \citep{Evert:2005}. Among the kept ones, we notice a prevalence of content words such as verbs, nouns and adjectives, whereas the typically pruned tokens are function words or word fragments.

\begin{figure*}[tb]
    \centering
    \begin{subfigure}{0.42\linewidth}
        \centering
        \includegraphics[width=\linewidth]{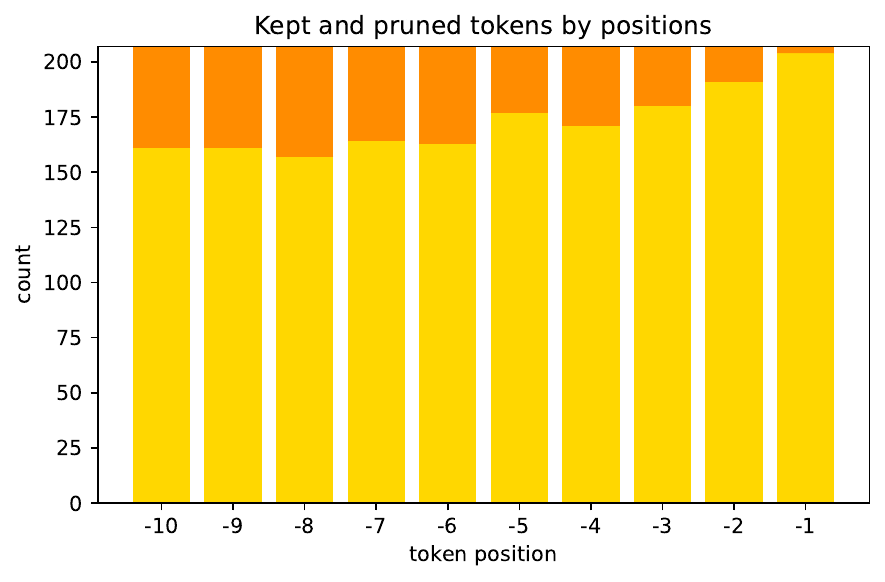}
        \captionsetup{aboveskip=0pt} %
        \caption{Autoprompts}
        \label{fig:kept_vs_pruned}
    \end{subfigure}
    \hfill
    \begin{subfigure}{0.42\linewidth}
        \centering
        \includegraphics[width=\linewidth]{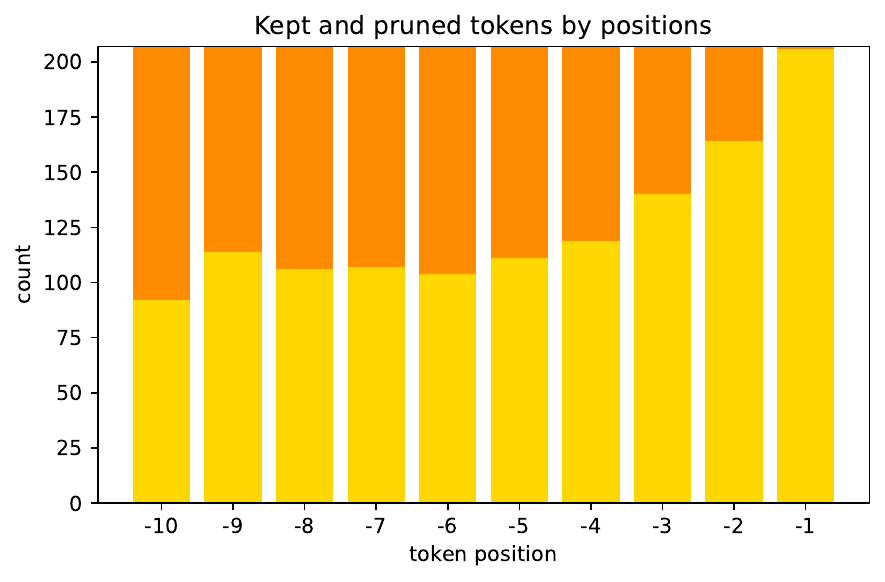}
        \captionsetup{aboveskip=0pt} %
        \caption{Original prompts (last 10 tokens)}
        \label{fig:kept_vs_pruned_nat}
    \end{subfigure}
    \caption{Counts of tokens that were pruned (dark orange) and kept (yellow) by position for Pythia-6.9B, where 0 is the last position. Tokens at last position are extremely unlikely to be pruned.}
    \label{fig:kept_vs_pruned_main}
\end{figure*}

\begin{figure}[t]
    \centering
    \includegraphics[width=0.95\linewidth]{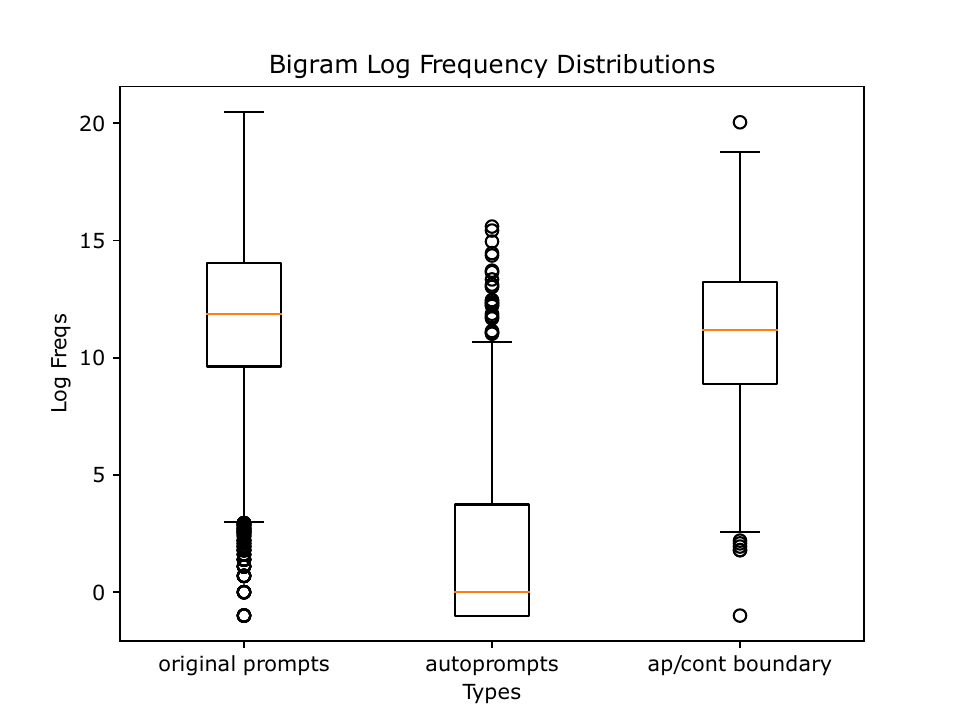}
    \caption{Pile-based log frequency distributions of bigrams in the \textit{original prompts}, \textit{autoprompts} and at the autoprompt/continuation boundary (\textit{ap/cont boundary}) for Pythia-6.9B. Log(0) conventionally set to -1. Red line = median; boxes span interquartile ranges.}
    \label{fig:bigram-freqs}
\end{figure}

\paragraph{Importance of the last token} The likelihood of pruning is not equally distributed across autoprompt positions: as Fig.~\ref{fig:kept_vs_pruned} shows, the \textit{last} token of the autoprompt is extremely unlikely to be pruned, pointing to the special role it plays in generating the continuation.\footnote{More generally, Fig.~\ref{fig:kept_vs_pruned} shows the last tokens before the very last also to be less prunable than earlier tokens.} By looking qualitatively at typical last tokens (see examples in Table~\ref{tab:merged-pruning-examples}), we observe indeed that often they have a natural link to the beginning of the continuation. To confirm this quantitatively, in Fig.~\ref{fig:bigram-freqs} we report the (log-transformed) corpus frequency distributions of the bigrams occurring in different contexts, with bigram frequencies estimated on the Pile corpus \citep{Gao:etal:2020} that was used to train the Pythia models.
There's a clear contrast between the bigram frequency distribution in natural text, exemplified by the natural prompts, and the autoprompts, that are mostly characterized by bigrams that never occur in the Pile. However, strikingly, the distribution at the autoprompt/continuation boundary is very similar to the one of natural text, quantitatively confirming that the last token of the autoprompt has a strong natural-language link to the continuation.

\subsection{Replacing autoprompt tokens}
\label{sec:replacing}

\begin{figure*}[tb]
    \centering
    \begin{subfigure}{0.40\linewidth}
        \centering
        \includegraphics[width=\linewidth]{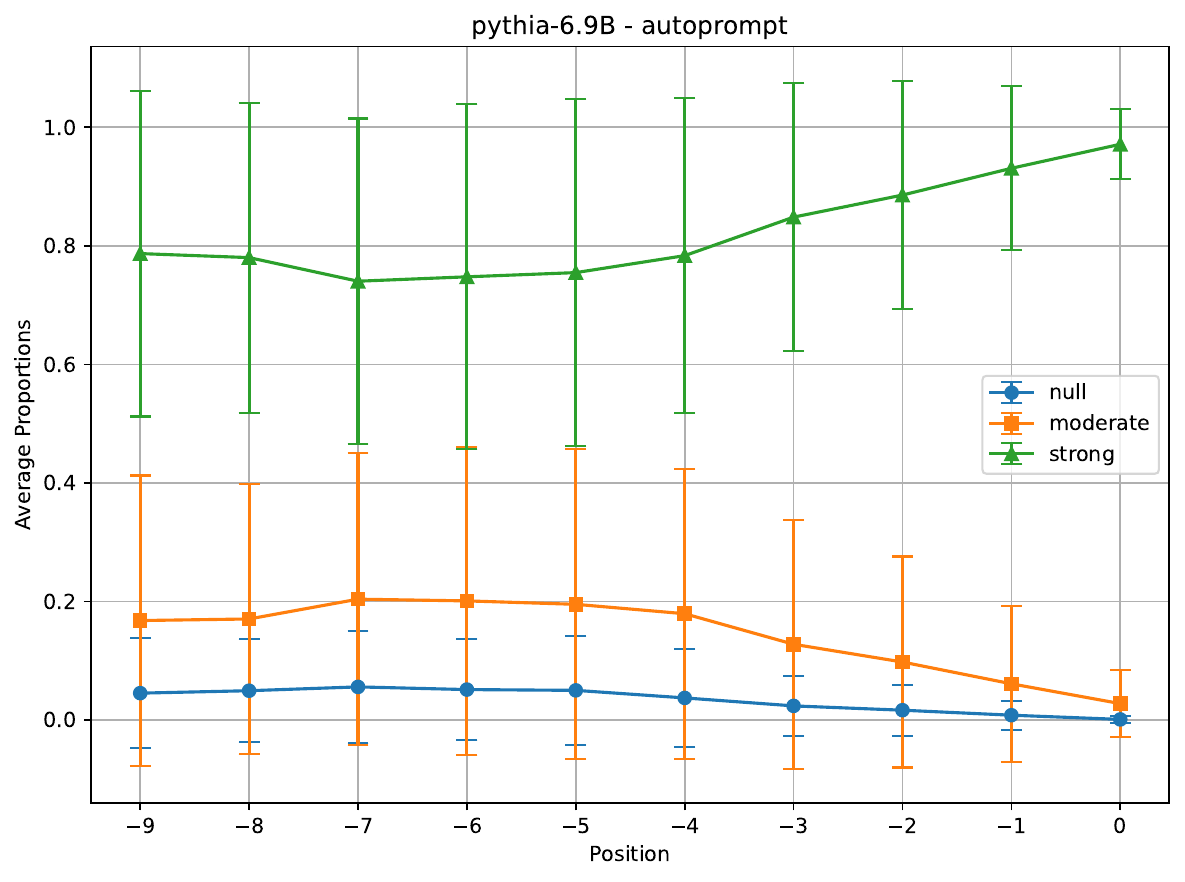}
        \captionsetup{aboveskip=0pt} %
        \caption{Autoprompts}
        \label{fig:replacement-distribution}
    \end{subfigure}
    \hfill
    \begin{subfigure}{0.40\linewidth}
        \centering
        \includegraphics[width=\linewidth]{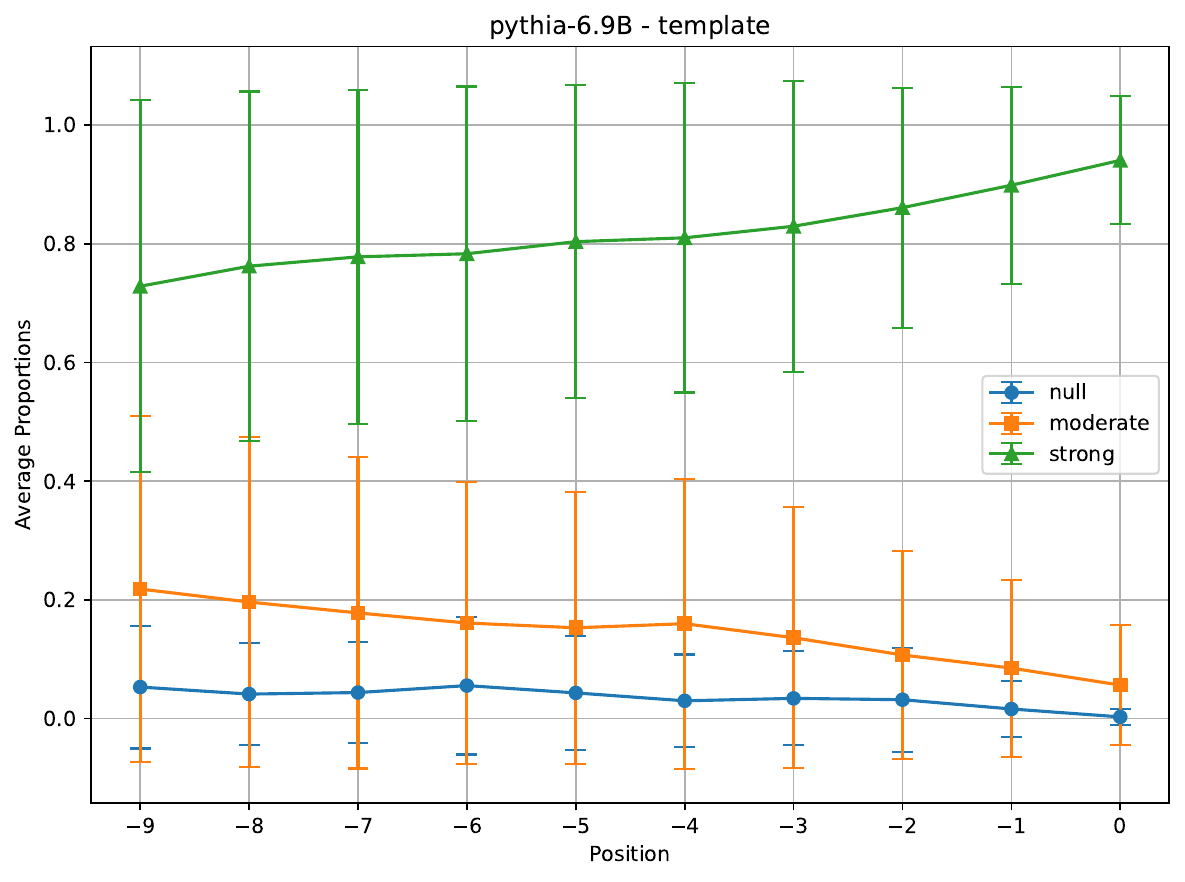}
        \captionsetup{aboveskip=0pt} %
        \caption{Original prompts (last 10 tokens)}
        \label{fig:replacement-distribution-original}
    \end{subfigure}
    \caption{Average proportions of replacement effect types by position on pruned prompts for Pythia-6.9B, aligned from right (whiskers show standard deviations). \textit{Null}-effect replacements leave the continuation unchanged. \textit{Moderate} replacements have BLEU $\geq 0.2$. \textit{Strong} replacements have BLEU $<0.2$.}
    \label{fig:replacement-distribution-main}
\end{figure*}

\begin{table}[t]
    \centering
    \begin{tabular}{lrrr}%
        \toprule
         & \textbf{All} & \textbf{Lang-like} & \textbf{Non-Ling} \\
        \midrule
        \textbf{Avg}  & 302.8 & 237.6 & 512.1 \\
        \textbf{Med}   &  15.0 &   9.0 &  73.5 \\
        \textbf{>50} & & 34\% & 55\% \\
        \bottomrule
    \end{tabular}
    \caption{Number of equivalent substitutes admitted by each autoprompt token for Pythia-6.9B. Results are shown for all, lang-like tokens, and non-ling tokens. Legend: \textbf{Avg}=average; \textbf{Med}=median; \textbf{>50}:\% of tokens with more than 50 equivalents.}
    \label{tab:rigidity}
\end{table}

\paragraph{Methodology} Working from now on with the pruned autoprompts, we replace the token in each position in turn with one of the 10k most frequent tokens from the Pile. We quantify the impact of the ablations in terms of BLEU score with respect to the original continuation. The ablation results are summarized in Fig.~\ref{fig:replacement-distribution}, where replacements are binned based on the impact they have on the continuation (examples are presented in App. \ref{app:replacement-examples}).

\paragraph{Impact of the replacements} First, we confirm that non-pruned tokens in all positions play a significant role in generating the continuation, as shown by the fact that most replacements have a \textit{strong} impact on BLEU. However, for all positions except the last, we also see that a non-negligible proportion of replacements do not affect the continuation at all, and in a significant proportion of cases the continuation is only mildly affected (as the examples in Table \ref{tab:replacement-examples-moderate} of App. \ref{app:replacement-examples} show, even a BLEU score $\approx$ 0.2 typically corresponds to a continuation that is quite similar to the original).
We confirm moreover the special role of the last token, that can almost never be replaced without catastrophic results. In general, as we approach the last position, it is increasingly more difficult to find replacements that do not strongly affect the continuation.

\arrayrulecolor{gray!50}
\begin{table*}[th]
    \centering
    \small  %
    \renewcommand{\arraystretch}{0.91}  %
    \begin{tabularx}{\linewidth}{p{7cm}X}
        \specialrule{1pt}{0pt}{0pt}
                \rowcolor{gray!20} \textbf{Autoprompts} & \textbf{Generated Continuation}   \\
        \specialrule{1pt}{0pt}{2pt}
        \firstbox{WHM} \firstbox{modelling}  
        \firstbox{tag}  
        \firstbox{Mus}  
        \secondbox{before}  
        \secondbox{either}  
        \secondbox{forming}  
        \secondbox{militant}  
        \firstbox{Annex}  
        \secondbox{Ireland}  
        & \dots or joining the Provisional IRA \\
        \noalign{\vskip 0.5ex}\hdashline\noalign{\vskip 0.5ex}
        \multicolumn{2}{l}{\textbf{before}: \textit{' then', ' change', ' subsequently'}} \\
        \multicolumn{2}{l}{\textbf{forming}: \textit{' meeting', ' leaving', ' remaining', ' developing', ' selling', ' breaking', ' producing', \dots}}\\
        \multicolumn{2}{l}{\textbf{Ireland}: \textit{' Irish'}}\\
        \multicolumn{2}{l}{\textbf{either/ militant}: \textit{0 substitutes.}} \\
        \midrule
        \firstbox{<|endoftext|>} \firstbox{<|endoftext|>} \firstbox{Star} \firstbox{Defense} \firstbox{*]\{\}}  
        \secondbox{1950}  
        \secondbox{blamed}  
        \secondbox{Tehran}  
        \secondbox{instead}  
        \secondbox{rhe}  
        & \dots United States for the 1953 coup. \\
        \noalign{\vskip 0.5ex}\hdashline\noalign{\vskip 0.5ex}
        \multicolumn{2}{l}{\textbf{1950}: \textit{' 1960'}}\\
        \multicolumn{2}{l}{\textbf{Tehran}: \textit{' Iran'}}\\
        \multicolumn{2}{l}{\textbf{instead}: \textit{' than', '=', ' then', ' back', ' to', '\&', ' again', ' though', \dots}}\\
        \multicolumn{2}{l}{\textbf{blamed/ rhe}: \textit{0 substitutes.}} \\
        \midrule
        \specialrule{1pt}{0pt}{0pt}
                \rowcolor{gray!20} \textbf{Original Prompt} & \textbf{Generated Continuation} \\
        \specialrule{1pt}{0pt}{2pt}
        \firstbox{for} 
        \secondbox{television} 
        \firstbox{Services} \firstbox{Our} \firstbox{is} 
        \secondbox{create} 
        \secondbox{programming} 
        \secondbox{that} 
        \secondbox{offers}  
        & \dots  a unique experience for our viewers. \\ %
        \noalign{\vskip 0.5ex}\hdashline\noalign{\vskip 0.5ex}
        \multicolumn{2}{l}{\textbf{television}: \textit{' Fox', ' entertainment'}}\\
        \multicolumn{2}{l}{\textbf{create}: \textit{' about',' emotional','Create','about',' differentiation',' unusual',' demands',' Create',\dots}}\\
        \multicolumn{2}{l}{\textbf{that}: \textit{'that'}}\\
        \multicolumn{2}{l}{\textbf{offers}: \textit{' provide',' provides',' offer'}}\\
        \multicolumn{2}{l}{\textbf{programming}: \textit{0 substitutes.}} \\
        \midrule
        \firstbox{Doctor} \firstbox{can} \firstbox{transcend} \firstbox{reach} 
        \secondbox{ep} 
        \secondbox{iph} 
        \secondbox{any} 
        \secondbox{and} 
        \secondbox{greater}
         & \dots understanding of the human condition. \\
         \noalign{\vskip 0.5ex}\hdashline\noalign{\vskip 0.5ex}
         \multicolumn{2}{l}{\textbf{ep}: \textit{' "',?,'**',' “',' new','“',' real',' story',']{}',' toward','ep',' towards','fn','Ñ', \dots}}\\
         \multicolumn{2}{l}{\textbf{iph}: \textit{'ph','rep'}}\\
         \multicolumn{2}{l}{\textbf{any}: \textit{' understanding','ening','aining',' insight'}}\\
         \multicolumn{2}{l}{\textbf{and}: \textit{' for',' by',' or','A',' \&',' through','or','an',' upon',' via','OR',' toward',' towards',' AND',' gain', \dots}}\\
         \multicolumn{2}{l}{\textbf{greater}: \textit{' improved',' enhanced',' deeper',' wider'}}\\
        \bottomrule
    \end{tabularx}
    \caption{Examples of replacement for autoprompts and original prompts (10 last tokens) for Pythia-6.9B. Color represents the number of substitutes: \firstbox{$>50$ substitutes} and \secondbox{$<50$ substitutes}. 
        When there are less than 50 substitutes, a random subset is displayed. Difficult-to-render characters are replaced by '?'. More examples in Table~\ref{tab:colored_sub_more} (appendix).}
    \label{tab:colored_sub}
\end{table*}
\arrayrulecolor{black}

\paragraph{\textit{Fillers} and \textit{key} tokens}
We further looked at the \textit{equivalent sets}, consisting, for a given token, of the substitutes that keep the continuation unchanged. We measure that $76.4\%$ of the tokens can be substituted by at least one equivalent, and each token has $302.8$ substitutes on average (cf. Table ~\ref{tab:rigidity}). But the size of the equivalent set is highly variable, with half of the tokens having $15.0$ or fewer equivalents.  This disparity is intuitively informative about the role of each token. In particular, we identify the presence of \textit{fillers} that can be replaced by a large number of substitutes, and \textit{key} tokens, that admit more restrained equivalent sets and must thus carry more specific information. This is evidenced by the fact that \textit{language-like} tokens tend to admit less equivalent substitutes than \textit{non-linguistic} tokens. %
Table~\ref{tab:colored_sub} (top section) shows examples of autoprompts with tokens color-coded based on their equivalent set size. When the equivalent set is small, we provide a random sample of it as an illustration.

\paragraph{Semantic consistency of the equivalent sets}
We measured the semantic consistency within equivalent sets in order to determine whether substitution in autoprompt is governed by semantic similarity (akin to synonymy in natural language) or by other factors. Specifically, we used FastText~\citep{Bojanowski:etal:2017} to compute the average semantic similarity: (a) between a token and each of its equivalents; and (b) among the equivalents themselves (including the original token).\footnote{For technical reasons, due to the difficulty of obtaining reliable representations for non-linguistic tokens, we focused only on language-like tokens.} The first measure indicates how semantically relevant the substitutes are in relation to the original token, while the second measure reflects the size of the substitute semantic space. A value close to $1$ suggests that the substitutes cluster within a small semantic region, meaning they are semantically consistent with one another.
We plot semantic similarity against set size in Fig~\ref{fig:sem_sub} (App.~\ref{app:sem}). As expected, small equivalent sets --corresponding to \textit{key} tokens-- tend to be semantically coherent on average. In this sense, they approximate near synonyms in natural language. This is confirmed by the examples in Table~\ref{tab:colored_sub}, e.g., `Ireland' is replaced by `Irish' and `Tehran' by `Iran'. Nonetheless, the semantic relation is often approximate, e.g., `before' being replaced by `then', `change' by `subsequently' or `forming' by `meeting', `leaving', `remaining', \textit{etc}. In contrast, large sets --associated to \textit{fillers}-- often include tokens that are semantically unrelated to each other. We show in Table~\ref{tab:dark_matter_full} (App.~\ref{app:replacement-examples}) the most frequently found tokens in large sets, mostly consisting of subwords (sometimes interpretable morphemes), digits and fragments of named entities that don't carry strong semantic information.

\paragraph{Can humans predict which autoprompt tokens are more important?} The four authors of the paper manually annotated the autoprompt dataset with a binary label marking, for each token, if it is intuitively important or not for the generation of the continuation. The annotation was made \textit{a posteriori}, using the full autoprompt and the generated continuation (see app.~\ref{app:annotation} for a description of the annotation process). Results show that autoprompts possess interpretable properties, as the labeling is correlated with the number of replacements a token might have. Indeed, we measured that the median size of the equivalent set for tokens deemed important is $2$, against $15$ for those deemed replaceable, with the effect observed both for language-like and non-linguistic tokens (Table~\ref{tab:human_prediction}).

\paragraph{Traces of compositionality} 
In cases where the replacement causes only a moderate change in the continuation, we see signs of ``compositionality,'' in the sense that the continuation only displays a few new tokens broadly reflecting the meaning of the replacement. Some examples are presented in Table \ref{tab:compositional-ap-examples}. We make the intuition more quantitative as follows. First, to facilitate automated similarity analysis, we extracted all cases where the replacement leads to the change of a single (typographic) word in the continuation (about 3\% of the total). For these cases, we used FastText to measure the semantic similarity of both the original autoprompt token and its replacement to the original word in the continuation and to the changed one. We found that the original token is more similar to the new continuation word (vs.~the original one) in only 48\% of the cases, whereas the replacement token is more similar to the new continuation in 59\% of the cases. We thus conclude that, indeed, there is a tendency for at least this type of replacement to work compositionally (a small change in the autoprompt leads to a semantically consistent change in the continuation). This, in turn, suggests that autoprompts do not function as unanalyzable holistic wholes. Their ``meaning'' to the model derives, at least partially, from assembling the meaning of its parts, as with natural language sequences. However, this looks nothing like the one performed by natural language syntax.

\arrayrulecolor{gray!50}
\begin{table*}[tb]
  \centering
  \resizebox{\textwidth}{!}{%
    \begin{tabular}{lr}
    \specialrule{1pt}{0pt}{0pt}
        \rowcolor{gray!20} \textbf{Autoprompts} & \textbf{Generated Continuation}   \\
    \specialrule{1pt}{0pt}{2pt}
       Eg<EOT> Brown \sout{mushrooms}/\textbf{face} chooses suffix "brown & \dots " to describe the color of the \sout{mushroom}/\textbf{face}.\\
        crossesFootnote Several panels accidentally have \sout{feather}/\textbf{bands} 517 chant collectively & \dots , as if they were a single \sout{bird}/\textbf{band}.\\
        fecture \sout{Phoenix}/\textbf{Toronto} Latinoamous]], effectively & \dots making it the largest city in \sout{Arizona}/\textbf{Canada}.\\
      \bottomrule
    \end{tabular}%
  }
  \caption{Example autoprompt token replacements leading to a small, interpretable change in the continuation for Pythia-6.9B (legend: \sout{replaced}/\textbf{replacement} in autoprompt; and \sout{original}/\textbf{new} in continuation).}
  \label{tab:compositional-ap-examples}
\end{table*}
\arrayrulecolor{black}

\subsection{Shuffling autoprompt tokens}
\label{sec:shuffling}

\paragraph{Methodology} Previous work has uncovered a somewhat mixed picture in terms of the robustness of autoprompts to token order shuffling \citep{Ishibashi:etal:2023,Rakotonirina:etal:2023,{Melamed:etal:2024}}. Based on our \textit{ad-interim} observations, we conjecture that the last token will be ``rigid'', as moving it around would strongly affect the continuation, whereas the preceding tokens might be more robust to order ablations. To test the conjecture, we randomly shuffled tokens (10 repetitions per autoprompt) and measured the resulting BLEU with respect to the original continuation. We either shuffled all tokens or left the last one fixed.

\paragraph{More than a bag-of-words} The average BLEU when shuffling all tokens is at 0.03 (s.d.~0.04) and at 0.06 (s.d.~0.11) when leaving the last token in its slot. This difference is highly significant (paired t-test, $p<0.001$). However, the low BLEU values suggest that, contrary to our conjecture, the autoprompt tokens before the last are not a bag of keywords, since their order matters as well. One possibility is that, while autoprompts as a whole do not constitute syntactically well-formed sequences, they are composed of tight sub-sequences that should not be separated. For example, given that modern tokenizers split text at the sub-word level, token-level shuffling will arbitrarily break words.
Some support for the view that the catastrophic effect of shuffling pre-last tokens is due to short-distance dependencies comes by looking at the cases in which a bigram in an autoprompt (excluding the last position) is also attested in the Pile corpus, either in the original or in the inverted order. In 60.5\% of these cases, the Pile frequency of the original bigram is larger than that of the inverted one, suggests some degree of natural local ordering among autoprompt tokens.

\subsection{Making human prompts more autoprompt-like}
\label{sec:natural-prompts}

As a final piece of evidence that the dynamics we see at work in autopromts are general properties of how LMs process language, we re-ran some of the experiments above on the original corpus-extracted natural-language prompts, finding that they respond in similar ways to our ablations.

\paragraph{Pruning} Applying the same greedy-pruning method to the original prompts, we find that more than 99.5\% can also be pruned, with 23.8 tokens removed on average (s.d.~13.2). Considering the average token length of the original prompts is 39.0, this means that, strikingly, on average 61\% of the tokens can be removed without affecting the continuation. Since the prompts are long, one could think that what is removed is primarily material towards the beginning of the sequence, but actually we find that 95\% of the prompts also have pruned tokens among the last 10 items. %
Examples of the latter are in Table \ref{tab:merged-pruning-examples} (bottom). Prunable material often consists of modifiers whose removal does not affect the basic syntactic structure of the fragments (``\textit{causing \textbf{the body's immune} cells}'', ``\textit{\textbf{Aviation} Regiment}''), but this is not always the case, and in many examples pruning turns well-formed sentences into seemingly unstructured token lists or telegraphic text at best. Still, like in the case of the autoprompts, the coherence of the transition between the prompt and the continuation is generally preserved (``\textit{\ldots a third / of which\ldots}'', ``\textit{\ldots as her / parents are\ldots}''). Table \ref{tab:lmi-merged} (right) shows the original-prompt tokens that are most typically kept vs.~pruned. As for the autoprompts, highly prunable tokens consist entirely of common function words and punctuation marks. However, typically kept tokens might also be (somewhat rarer) function words and punctuation marks. %
Figure~\ref{fig:kept_vs_pruned_nat} presents pruning proportion by position for the last 10 tokens in the original prompts, confirming that, in this case as well, the last token is by far the most important one in determining the continuation. Interestingly, the contrast is even more dramatic than for autoprompts (Figure~\ref{fig:kept_vs_pruned}).

\paragraph{Replacement} We replicate the token-replacement experiment on the pruned original prompts, obtaining the results summarized in Figure \ref{fig:replacement-distribution-original}. Again, tokens become more replaceable as we move away from the end of the prompt, confirming the crucial role played by the very last token. Moreover, we confirm the presence of both \textit{key} tokens, having few semantically related substitutes, and \textit{fillers} with numerous semantically inconsistent  substitutes (Figure~\ref{fig:sem_sub_template}).
Table \ref{tab:colored_sub} (bottom) shows examples in which the original prompt, despite pruning and replacement among the last 10 tokens, still triggers the same continuation. We see how the same principles that might explain the success of autoprompts are at work here, suggesting how autoprompts might take shape during the induction process.

\paragraph{Shuffling} Shuffling all tokens of the original prompts after pruning leads to an average BLEU of 0.02 (s.d.~0.02), comparably to autoprompts. Leaving the last token in place leads to an average BLEU of 0.03 (s.d.~0.05). This small difference is highly significant (paired t-test, $p<0.001$), confirming the importance of the last token (the difference stays equally significant if we compare shuffling all but the last token to shuffling while keeping one random non-last token fixed).

\section{Discussion}

Our findings about autoprompts, confirmed by autoprompt-inspired ablations of natural prompts, suggest that LMs might rely on a simplified model of language, where not all tokens have specific syntactic and semantic functions in an abstract syntactic tree. We note that the phenomenon of relying on over-simplified representations of the data is not specific to LMs. Convolutional Neural Network classifiers of visual data also latch onto superficial correlations in the data, leading to poor ood generalization \citep{jo2017measuring,ilyas2019adversarial,yin2019fourier,geirhos2020shortcut}. 

While we hope our results are of general interest, we recognize a number of limitations. First, due to the time it takes to induce autoprompts with our computational resources, we could only experiment with 6 models, the largest of which has 12B parameters. We make our code available in hope that researchers with bigger resources will run similar experiments on a larger scale. For analogous reasons, we only experimented with one variant of the autoprompt inducing algorithm.
Given that all algorithms we are aware of adopt similar gradient-following methods, and based on qualitative inspection of autoprompt examples in other papers, we expect our conclusions to hold for autoprompts independently of how they are induced, but this should be verified empirically.

\section*{Acknowledgments}
We thank Emmanuel Chemla, Mor Geva and audiences at SISA, CIMeC and at the online HiTZ seminar series for feedback. Our work was funded by the European Research Council (ERC) under the European Union’s Horizon 2020 research and innovation programme (grant agreement No. 101019291).  We also received funding from  the Catalan government (AGAUR grant SGR 2021 00470).  This paper reflects the authors’ view only, and the funding agencies are not responsible for any use that may be made of the information it contains.

\bibliography{marco,custom}

\newpage

\appendix

\section*{Ethics Statement}
If we do not achieve a genuine understanding of how LMs process and generate text, we cannot fully control their behavior and mitigate unintended or intentional harm. Opaque autoprompts are an indication that there are important aspects of LM prompting and generation that are still out of our control. Our investigation into the nature of this phenomenon contributes to a better understanding of how LMs work and, thus, ultimately, to make them safer and more predictable.

\section{Results with other models}
\label{app:other-models}
\paragraph{Data-set statistics} Following the procedure described in Section \ref{sec:methods}, we build a data-set for each of the following models: Pythia-1.4B, OLMo-1B, OLMo-7B, OLMo-7B-Instruct. Data-set statistics are presented in Table \ref{tab:dataset_statistics}. 

\begin{table*}
    \centering
    \begin{tabular}{lrrrrr}
    \toprule
       \textbf{Model} &  \textbf{Data-set size} & \textbf{Original prompt length} & \textbf{Continuation length} \\
    \midrule
         Pythia-1.4B & 2473 & 38.6 (11.7) & 9.4 (2.7) \\
         Pythia-12B & 129 & 37.5 (11.0) & 7.9 (1.7) \\
         OLMo-1B & 500 & 38.4 (11.2) & 8.5 (2.0) \\
         OLMo-7B & 115 & 38.9 (10.9) & 8.3 (1.9) \\
         OLMo-7B-Instruct & 104 & 39.8(12.7) & 8.4(2.7) \\
    \bottomrule 
    \end{tabular}
    \caption{The number of entries, average original prompt length (s.d.), and average continuation length (s.d.) of the additional data-sets.}
    \label{tab:dataset_statistics}
\end{table*}

\paragraph{Pruning and shuffling} The results of the autoprompt pruning and shuffling experiments are presented in Table \ref{tab:autoprompt_experiments}. For all models, there is a difference in BLEU when shuffling all tokens vs. keeping last token fixed (paired t-test significant at p < 0.01). The difference stays comparably significant if, in the first condition, we leave a random non-last token fixed, so that the same number of tokens is shuffled in the two cases. Token pruning distribution by position is shown in Figures~\ref{fig:exp_pythia1.3b}~\ref{fig:exp_olmo1b}~\ref{fig:exp_olmo7b}~\ref{fig:exp_pythia12b}~\ref{fig:exp_olmo7b_inst} (a).

\begin{table*}
    \centering
    \begin{tabular}{lrrrrr}
    \toprule
       \textbf{Model} &  \textbf{Pruning rate} & \begin{tabular}[c]{@{}l@{}}\textbf{Tokens}\\ \textbf{pruned}\end{tabular} & \begin{tabular}[c]{@{}l@{}}\textbf{BLEU}\\ \textbf{shuffle all}\end{tabular} & \begin{tabular}[c]{@{}l@{}}\textbf{BLEU}\\ \textbf{shuffle except last}\end{tabular} \\
    \midrule
         Pythia-1.4B & 60\% & 1.2 (1.3) & 0.03 (0.03) & 0.05 (0.07) \\
         Pythia-12B & 86\% & 2.9 (1.9) & 0.04 (0.05) & 0.09 (0.12) \\
         OLMo-1B & 60\% & 1.2 (1.3) & 0.02 (0.01) & 0.04 (0.04) \\
         OLMo-7B & 60\% & 1.2 (1.4) & 0.02 (0.02) & 0.04 (0.05) \\
         OLMo-7B-Instruct & 27\% & 0.4(0.7) & 0.01 (0.01) & 0.02 (0.03) \\
    \bottomrule 
    \end{tabular}
    \caption{Results of the pruning and shuffling experiments of autoprompts. Pruning rate is the proportion of prompts in which at least one token could be  pruned. There is a difference between BLEU scores when shuffling all tokens vs.~keeping last token fixed for all models (paired t-test significant at p < 0.01).}
    \label{tab:autoprompt_experiments}
\end{table*}

\paragraph{Replacement} For OLMo models, we estimate the top 10k most frequent tokens to be used in the replacement experiments using a sample of approximately 10 billion tokens from the Dolma corpus, which was used to train this model. \citep{Soldaini:etal:2024}. Proportions of replacement effect type by position are reported in Figures~\ref{fig:exp_pythia1.3b}~\ref{fig:exp_olmo1b}~\ref{fig:exp_olmo7b}~\ref{fig:exp_pythia12b}~\ref{fig:exp_olmo7b_inst} (b).

\paragraph{Semantic consistency} Similarly, we measure the semantic consistency of the equivalent sets in Figures~\ref{fig:exp_pythia1.3b}~\ref{fig:exp_olmo1b}~\ref{fig:exp_olmo7b}~\ref{fig:exp_pythia12b}~\ref{fig:exp_olmo7b_inst} (c).

\subsection{Making prompts more autoprompt-like}

\begin{table*}
\small
    \centering
    \begin{tabular}{lrrrrr}
    \toprule
       \textbf{Model} &  \textbf{Pruning rate} & \begin{tabular}[c]{@{}l@{}}\textbf{Pruning rate}\\ \textbf{last 10 tokens}\end{tabular} & \begin{tabular}[c]{@{}l@{}}\textbf{Tokens}\\ \textbf{pruned}\end{tabular} & \begin{tabular}[c]{@{}l@{}}\textbf{BLEU}\\ \textbf{shuffle all}\end{tabular} & \begin{tabular}[c]{@{}l@{}}\textbf{BLEU}\\ \textbf{shuffle except last}\end{tabular} \\
    \midrule
         Pythia-1.4B & 99\% & 95\% & 21.9(12.3) & 0.02 (0.03) & 0.03 (0.05) \\
         Pythia-12B & 100\% & 99\% & 23.8(11.7) & 0.03 (0.04) & 0.05 (0.12) \\
         OLMo-1B & 100\% & 97\% & 23.4(12.3) & 0.02 (0.03) & 0.03 (0.05) \\
         OLMo-7B & 100\% & 98\% & 24.0(12.6) & 0.02 (0.02) & 0.04 (0.06) \\
         OLMo-7B-Instruct & 92\% & 90\% & 21.4(14.6) & 0.01 (0.01) & 0.02 (0.03) \\
    \bottomrule 
    \end{tabular}
    \caption{Results of the pruning and shuffling experiments of original prompts. There is a difference between BLEU scores when shuffling all tokens vs. keeping last token fixed for all models except OLMo-7B-Instruct  (paired t-test significant at p < 0.01).}
    \label{tab:natural_prompt_experiments}
\end{table*}

\begin{figure*}[t]
    \centering
    \begin{subfigure}{0.45\linewidth}
        \centering
        \includegraphics[width=\linewidth]{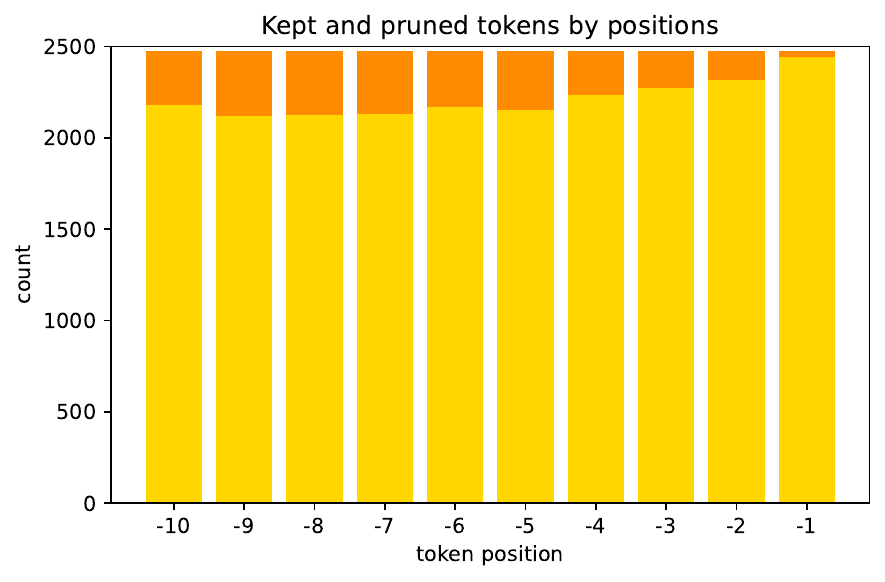}
    \end{subfigure}
    \hfill
    \begin{subfigure}{0.45\linewidth}
        \centering
        \includegraphics[width=\linewidth]{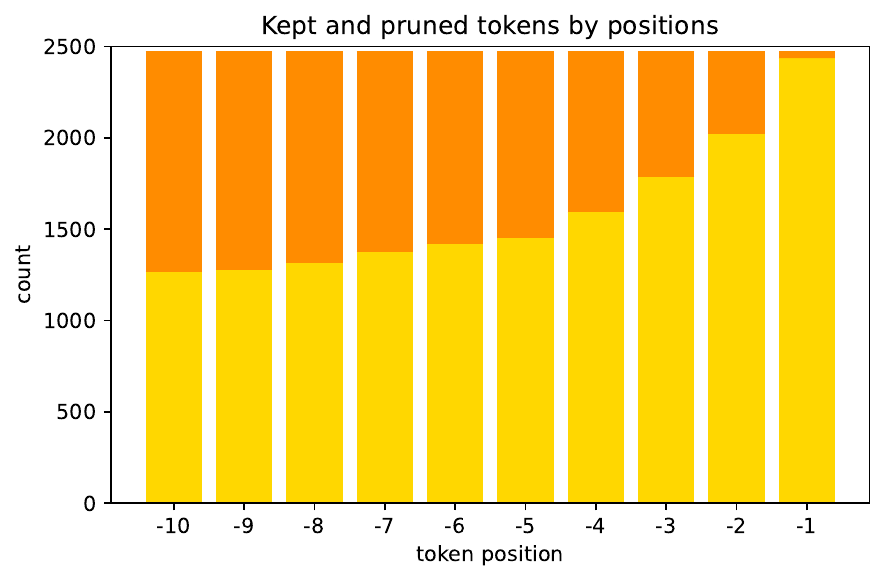}
    \end{subfigure}
    \caption*{(a) Count of pruned (orange) and kept (yellow) tokens (\textit{left:} autoprompt; \textit{right:} original prompt).}

    \begin{subfigure}{0.45\linewidth}
        \centering
        \includegraphics[width=\linewidth]{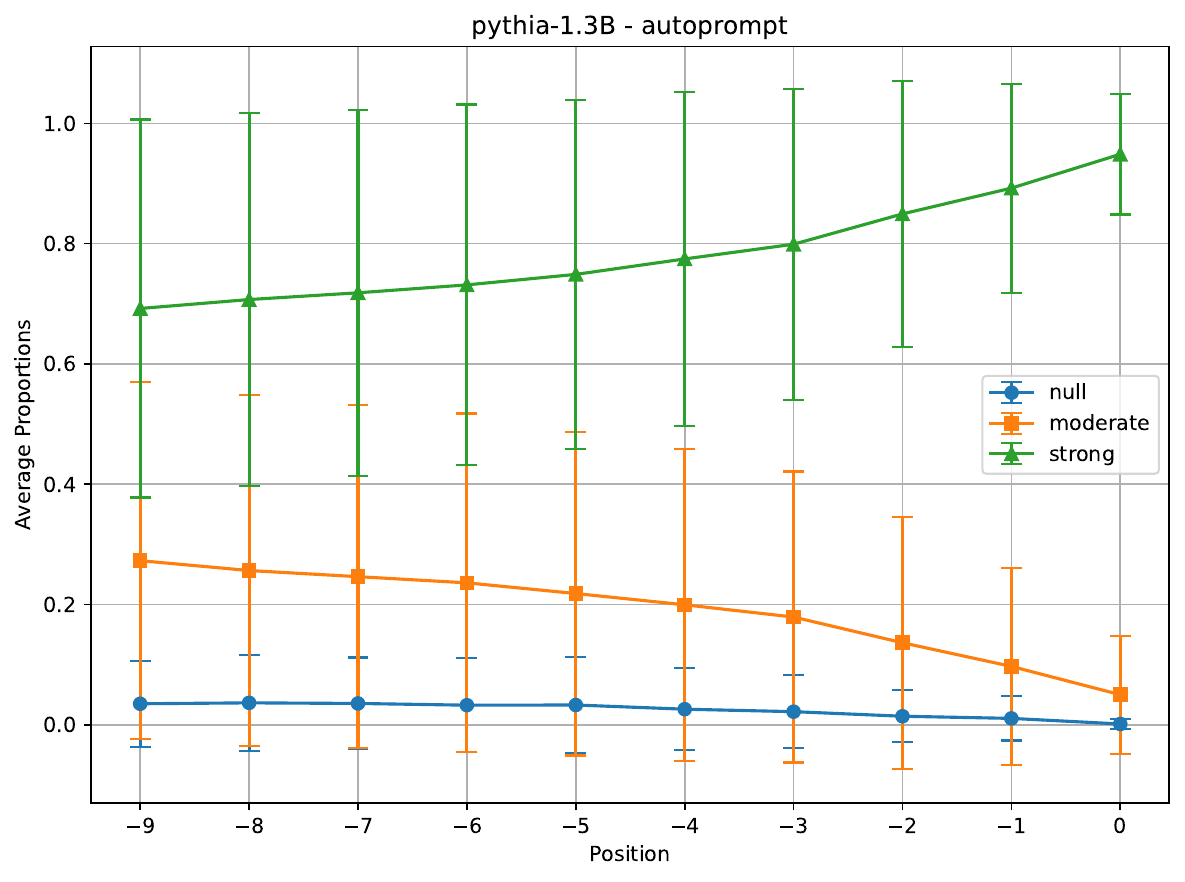}
    \end{subfigure}
    \hfill
    \begin{subfigure}{0.45\linewidth}
        \centering
        \includegraphics[width=\linewidth]{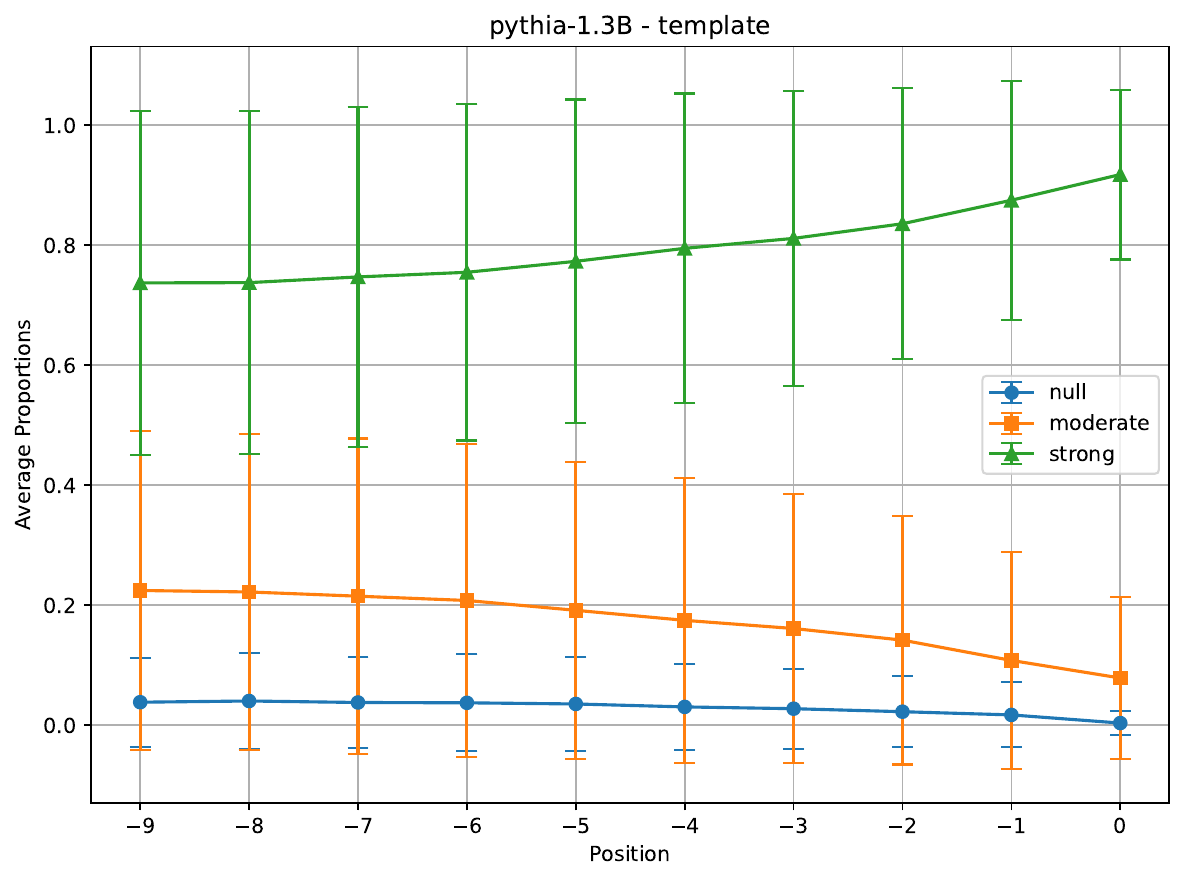}
    \end{subfigure}
    \caption*{(b) Average proportions of replacement effect types by position (\textit{left:} autoprompt; \textit{right:} original prompt).}

    \begin{subfigure}{0.45\linewidth}
        \centering
        \includegraphics[width=\linewidth]{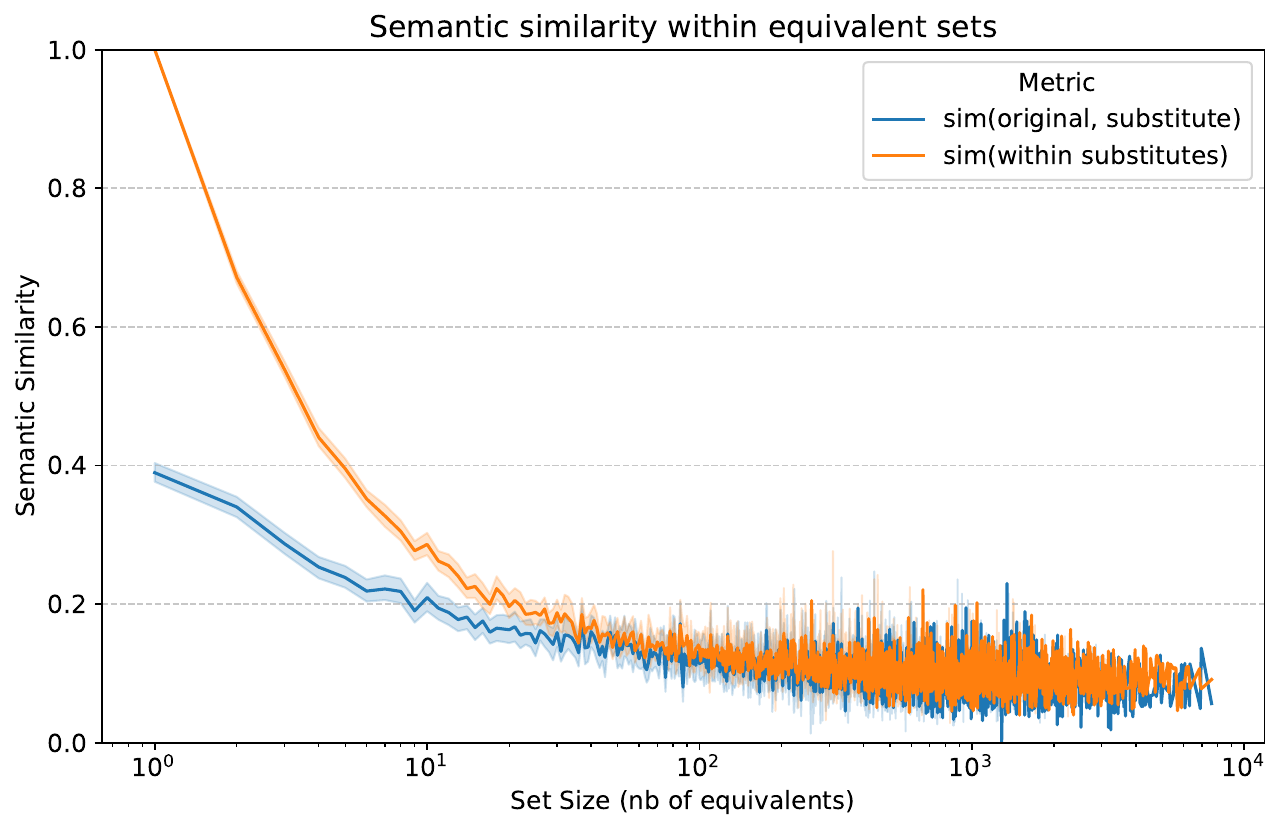}
    \end{subfigure}
    \hfill
    \begin{subfigure}{0.45\linewidth}
        \centering
        \includegraphics[width=\linewidth]{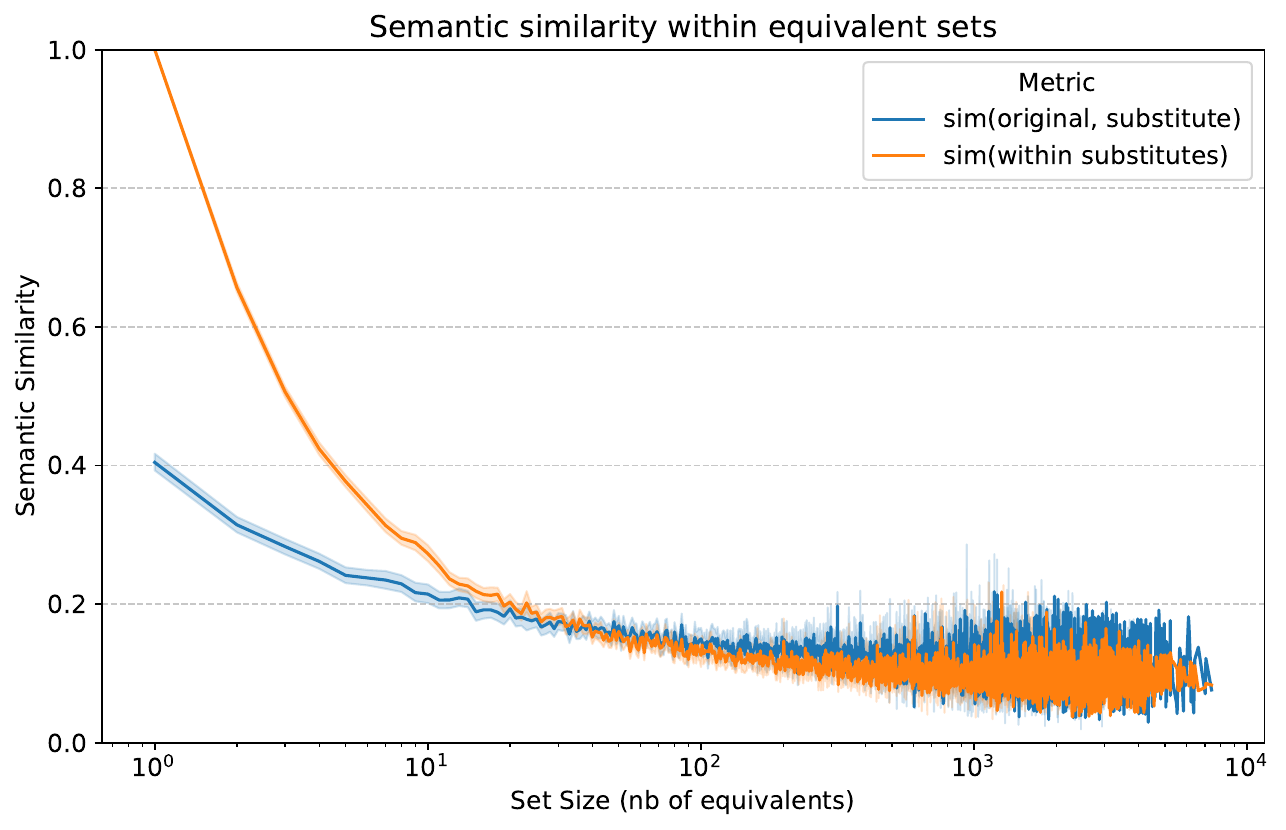}
    \end{subfigure}
    \caption*{(c) Semantic similarity between substitutes (\textit{left:} autoprompt; \textit{right:} original prompt).}

    \caption{\textbf{Pythia-1.3B}: Reproducing pruning and replacement experiments.}
    \label{fig:exp_pythia1.3b}
\end{figure*}

\begin{figure*}[t]
    \centering
    \begin{subfigure}{0.45\linewidth}
        \centering
        \includegraphics[width=\linewidth]{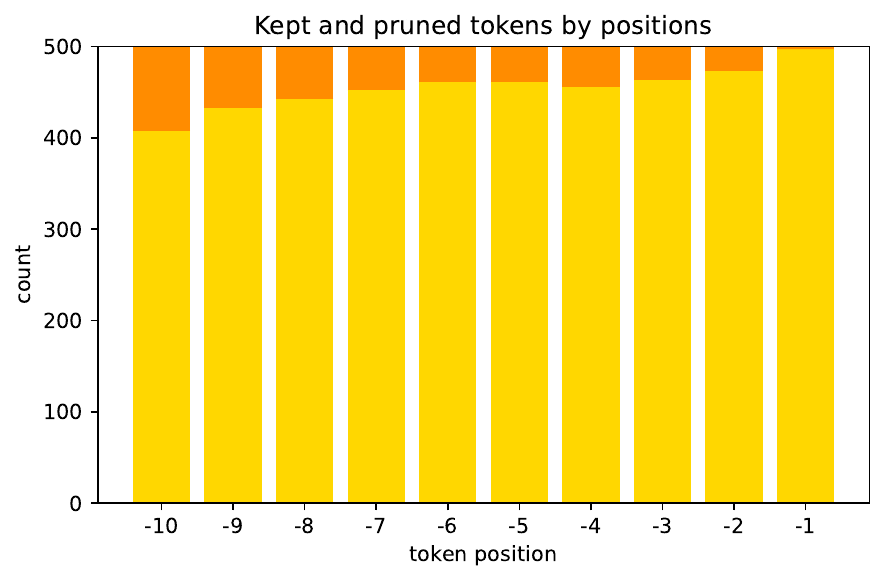}
    \end{subfigure}
    \hfill
    \begin{subfigure}{0.45\linewidth}
        \centering
        \includegraphics[width=\linewidth]{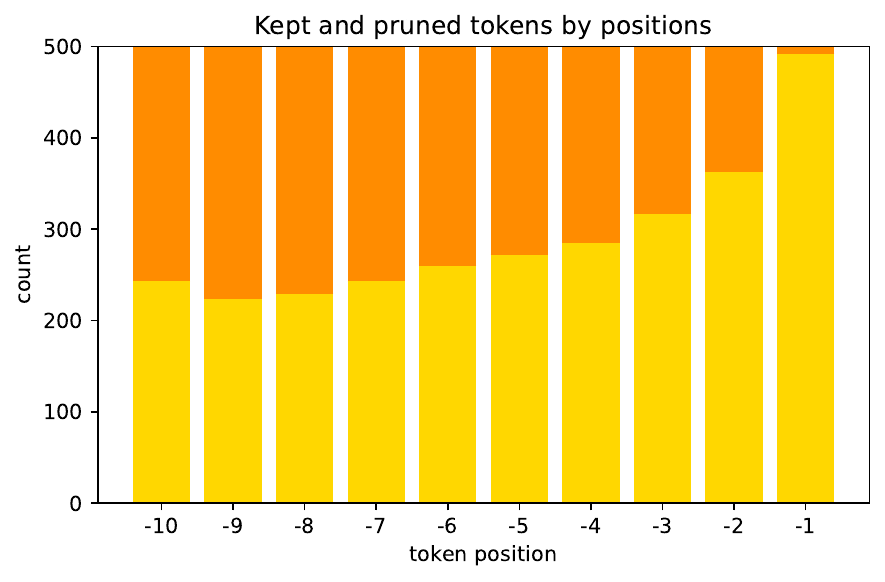}
    \end{subfigure}
    \caption*{(a) Count of pruned (orange) and kept (yellow) tokens (\textit{left:} autoprompt; \textit{right:} original prompt).}

    \begin{subfigure}{0.45\linewidth}
        \centering
        \includegraphics[width=\linewidth]{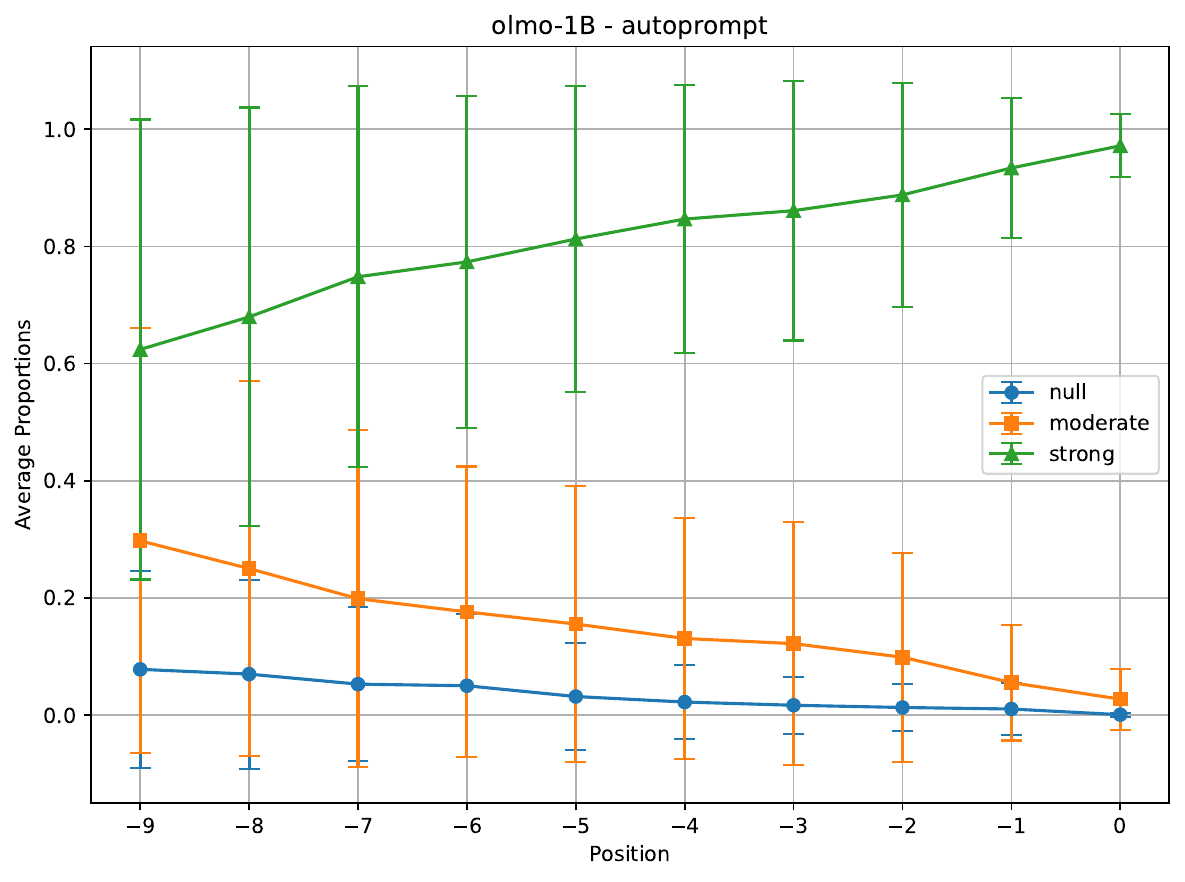}
    \end{subfigure}
    \hfill
    \begin{subfigure}{0.45\linewidth}
        \centering
        \includegraphics[width=\linewidth]{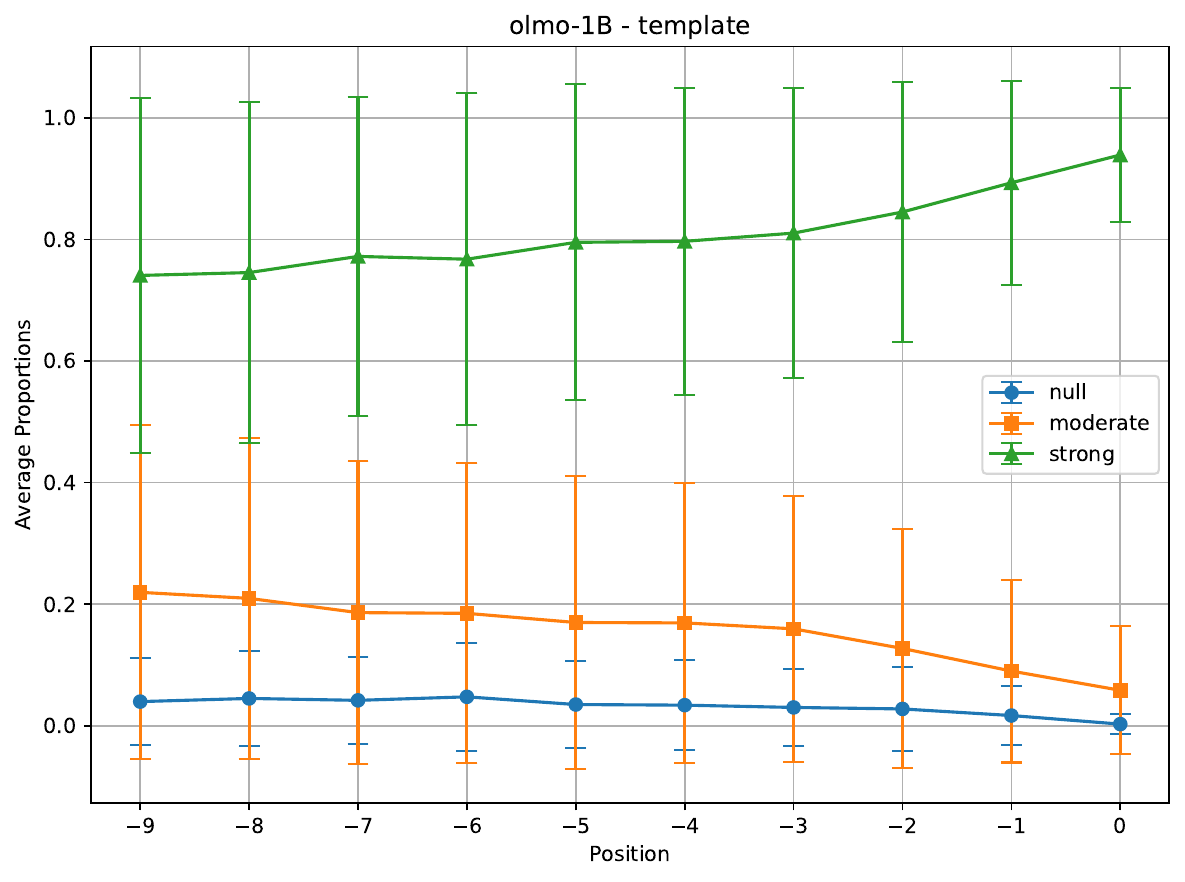}
    \end{subfigure}
    \caption*{(b) Average proportions of replacement effect types by position (\textit{left:} autoprompt; \textit{right:} original prompt).}

    \begin{subfigure}{0.45\linewidth}
        \centering
        \includegraphics[width=\linewidth]{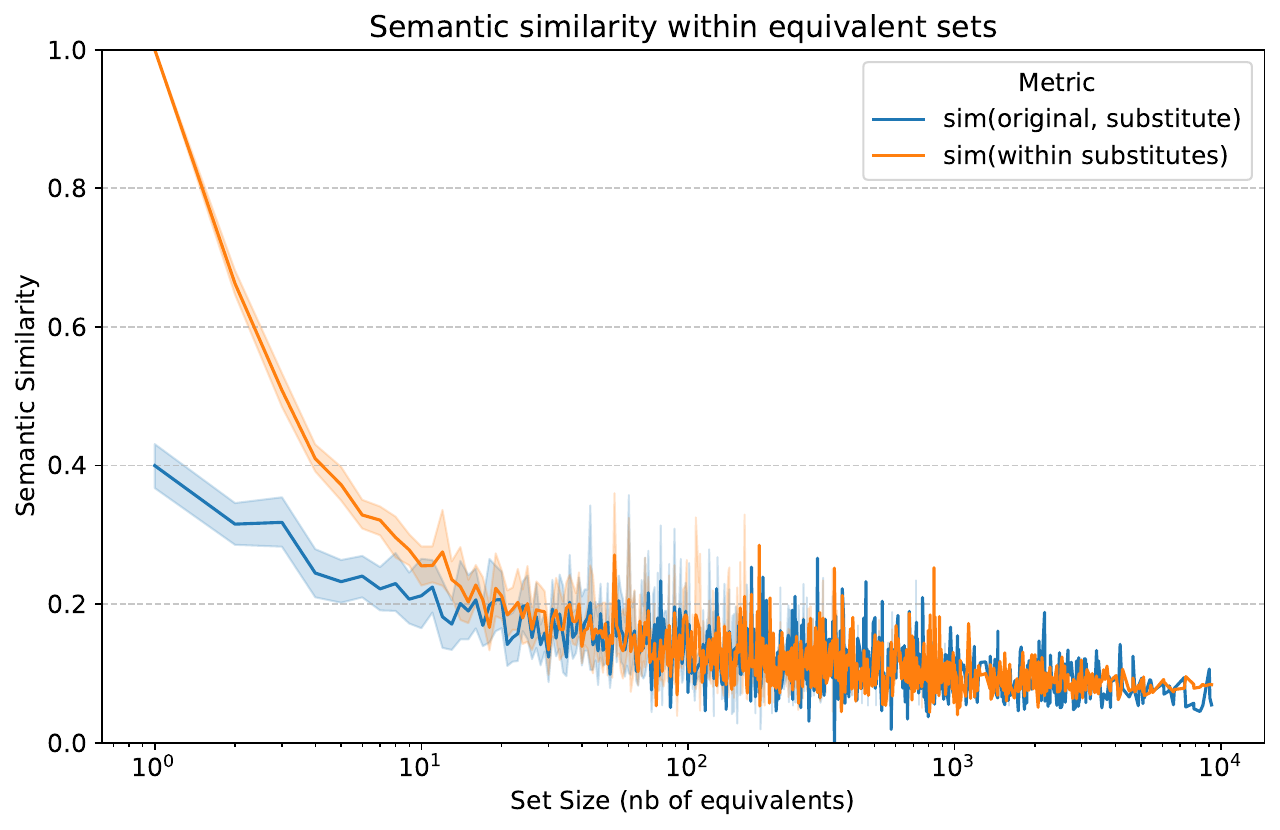}
    \end{subfigure}
    \hfill
    \begin{subfigure}{0.45\linewidth}
        \centering
        \includegraphics[width=\linewidth]{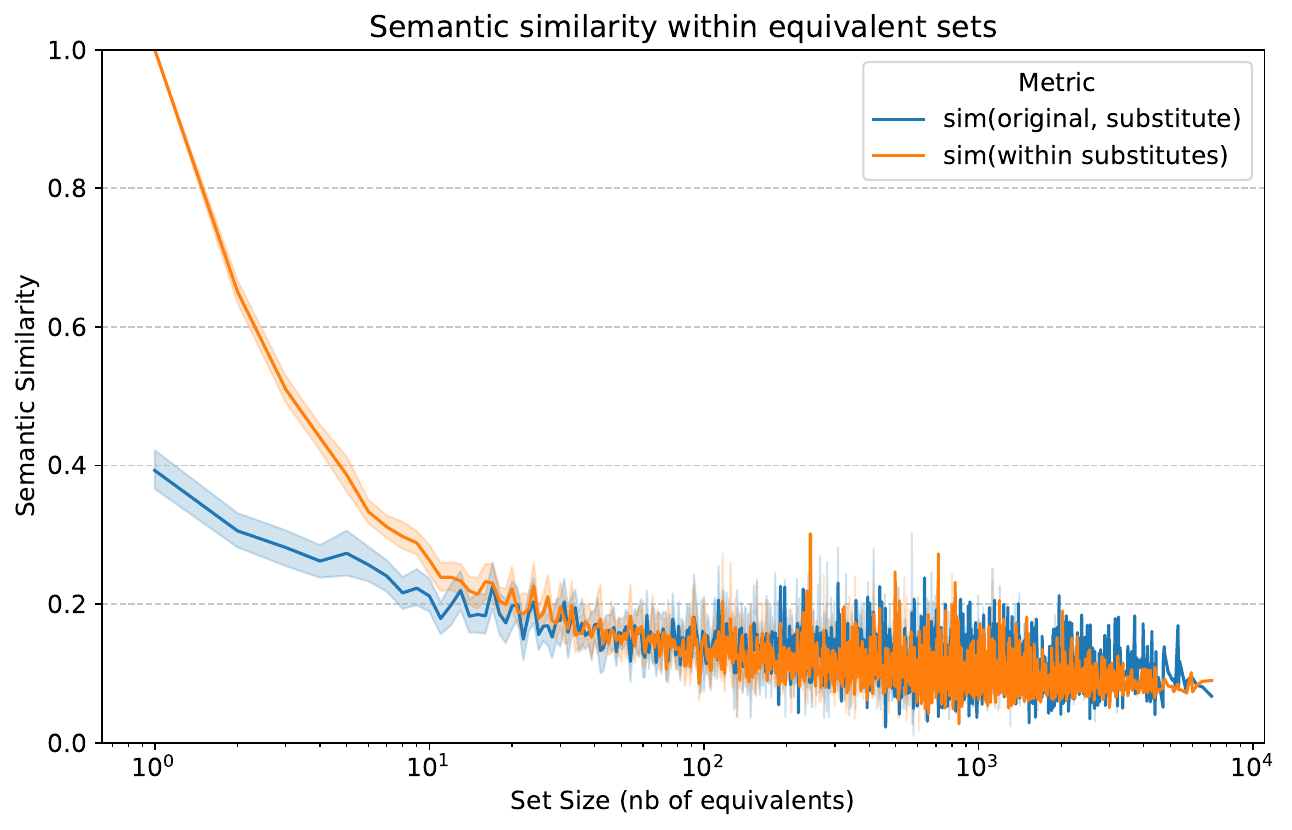}
    \end{subfigure}
    \caption*{(c) Semantic similarity between substitutes (\textit{left:} autoprompt; \textit{right:} original prompt).}

    \caption{\textbf{OLMo-1B}: Reproducing pruning and replacement experiments.}
    \label{fig:exp_olmo1b}
\end{figure*}

\begin{figure*}[t]
    \centering
    \begin{subfigure}{0.45\linewidth}
        \centering
        \includegraphics[width=\linewidth]{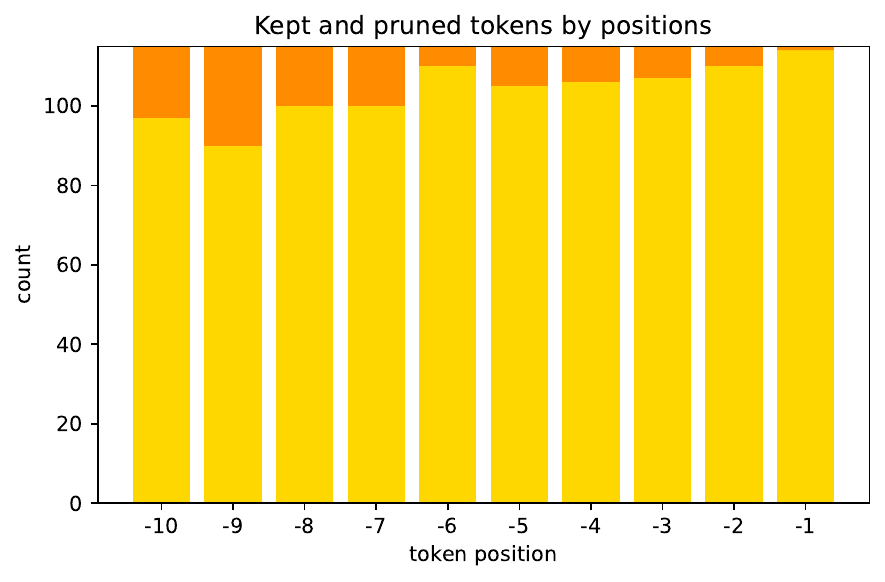}
    \end{subfigure}
    \hfill
    \begin{subfigure}{0.45\linewidth}
        \centering
        \includegraphics[width=\linewidth]{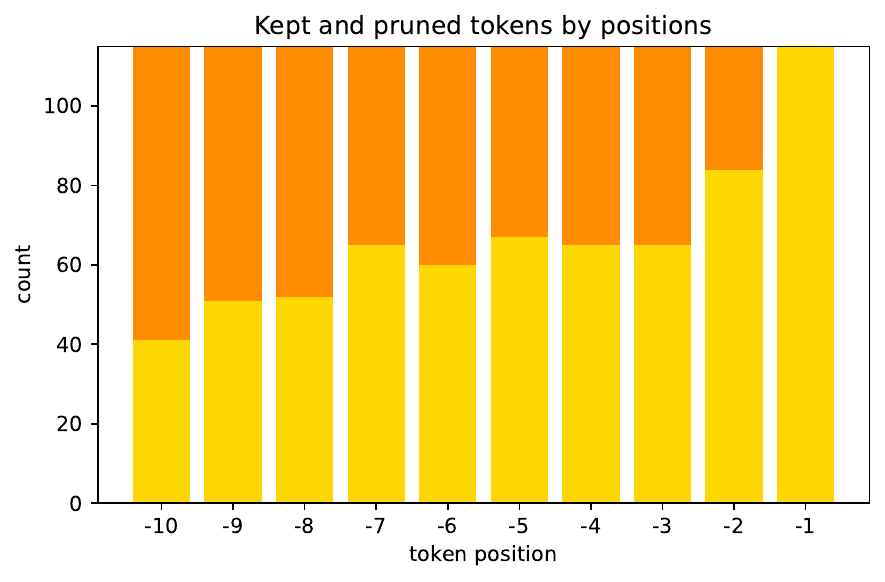}
    \end{subfigure}
    \caption*{(a) Count of pruned (orange) and kept (yellow) tokens (\textit{left:} autoprompt; \textit{right:} original prompt).}

    \begin{subfigure}{0.45\linewidth}
        \centering
        \includegraphics[width=\linewidth]{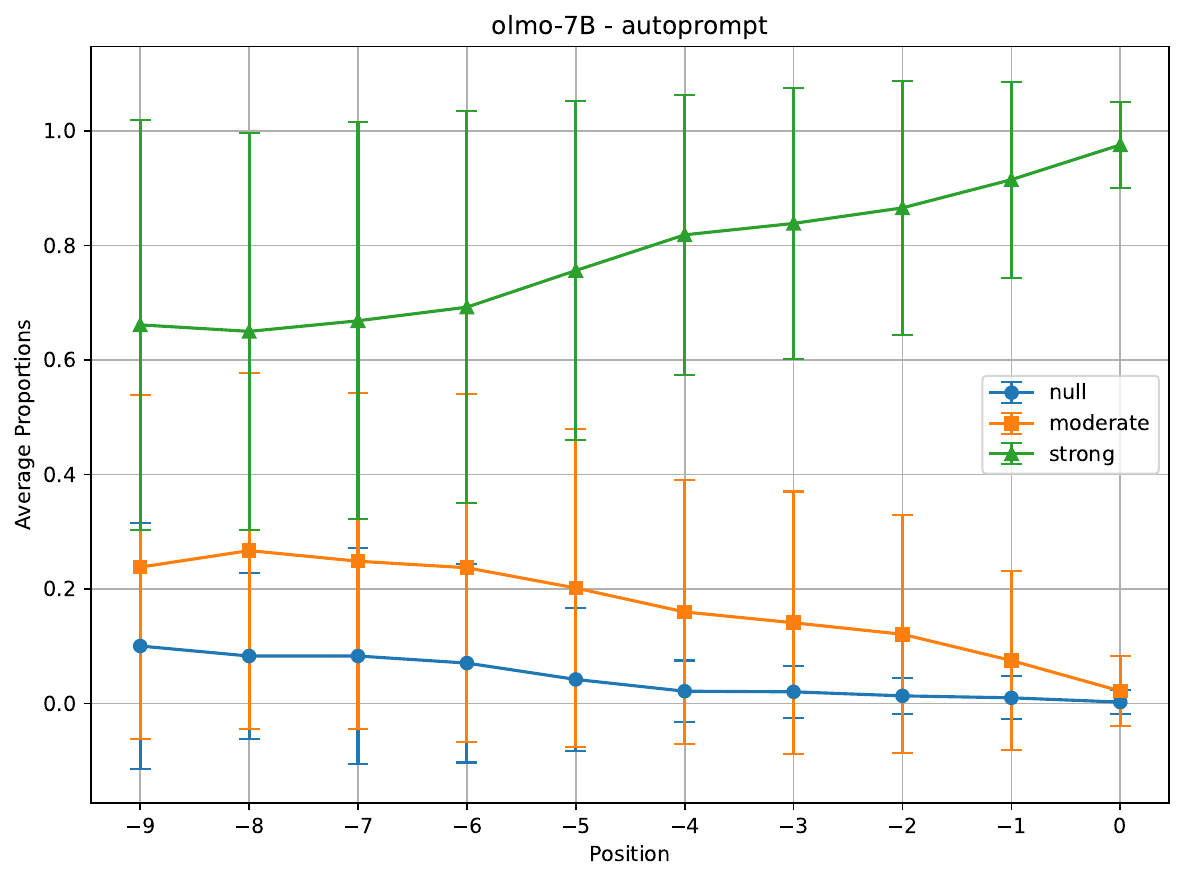}
    \end{subfigure}
    \hfill
    \begin{subfigure}{0.45\linewidth}
        \centering
        \includegraphics[width=\linewidth]{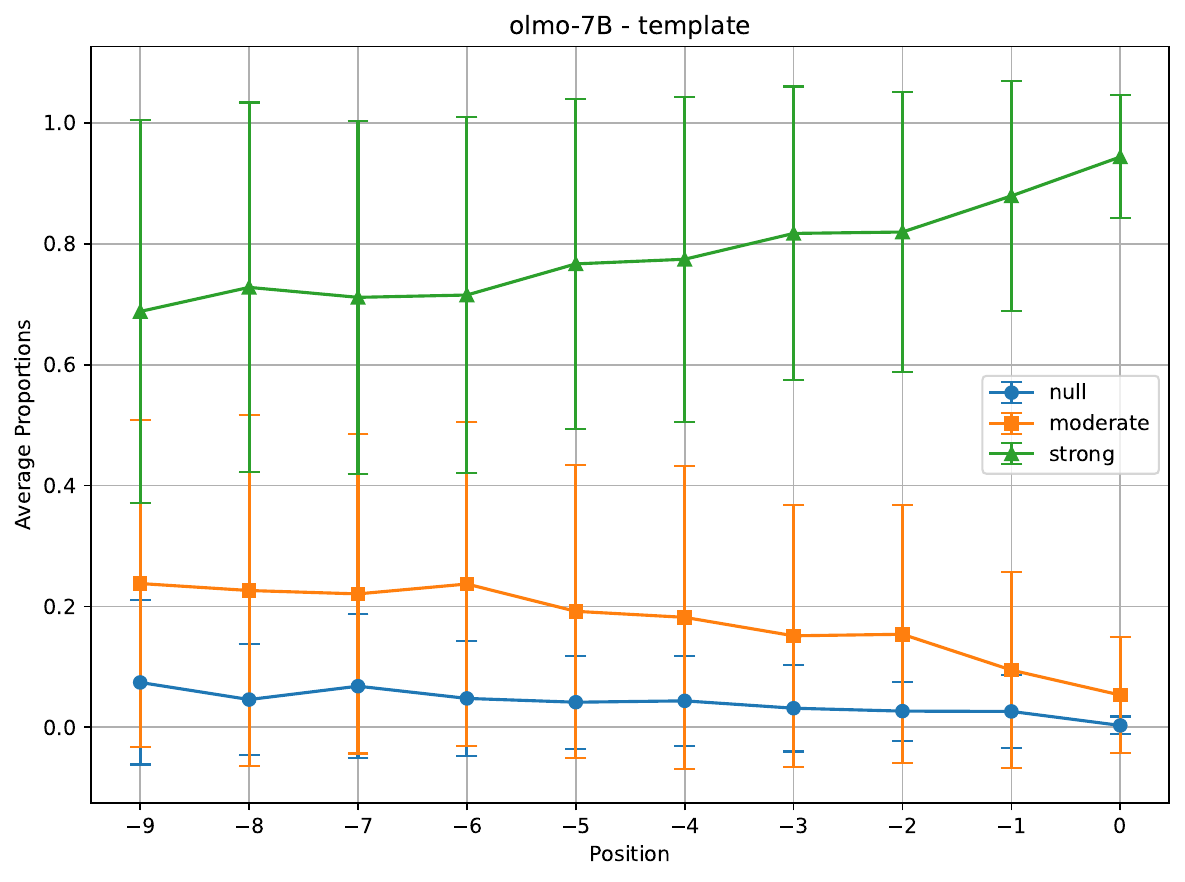}
    \end{subfigure}
    \caption*{(b) Average proportions of replacement effect types by position (\textit{left:} autoprompt; \textit{right:} original prompt).}

    \begin{subfigure}{0.45\linewidth}
        \centering
        \includegraphics[width=\linewidth]{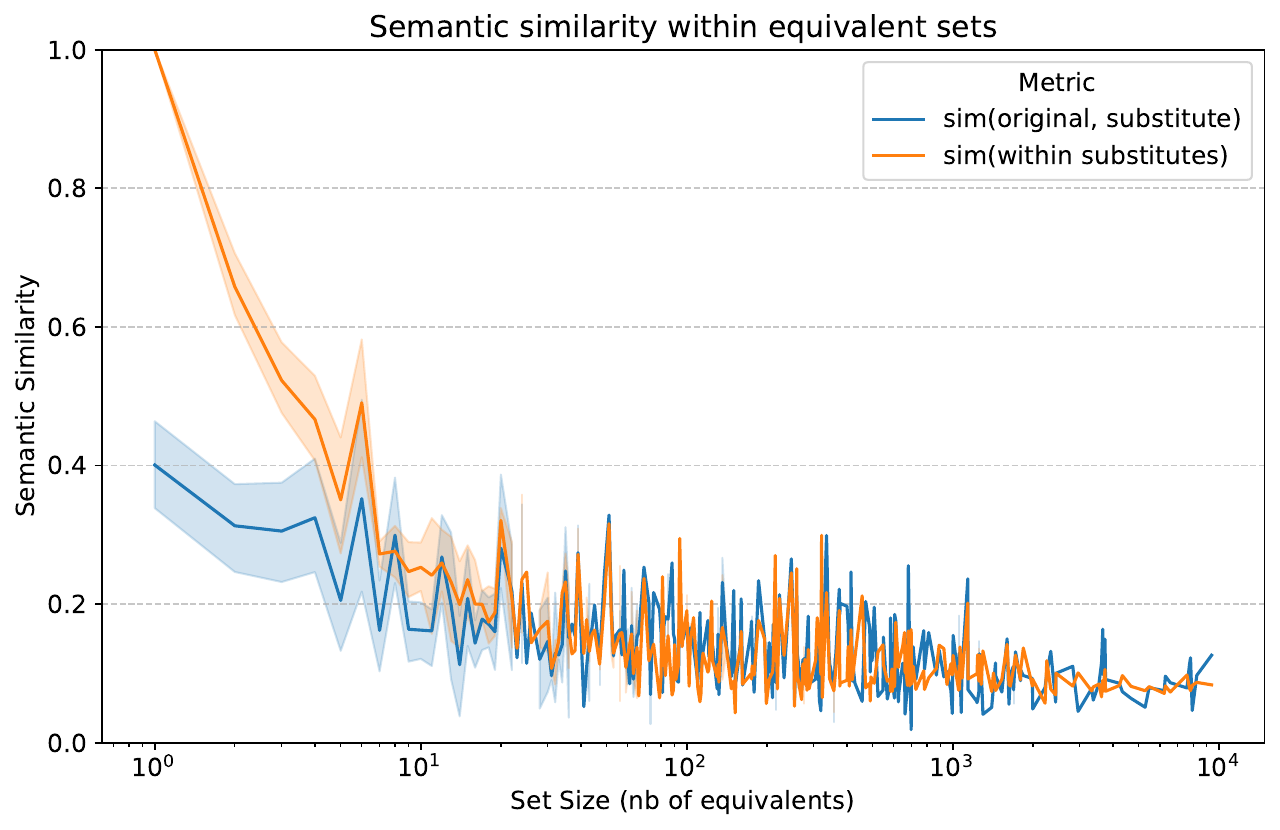}
    \end{subfigure}
    \hfill
    \begin{subfigure}{0.45\linewidth}
        \centering
        \includegraphics[width=\linewidth]{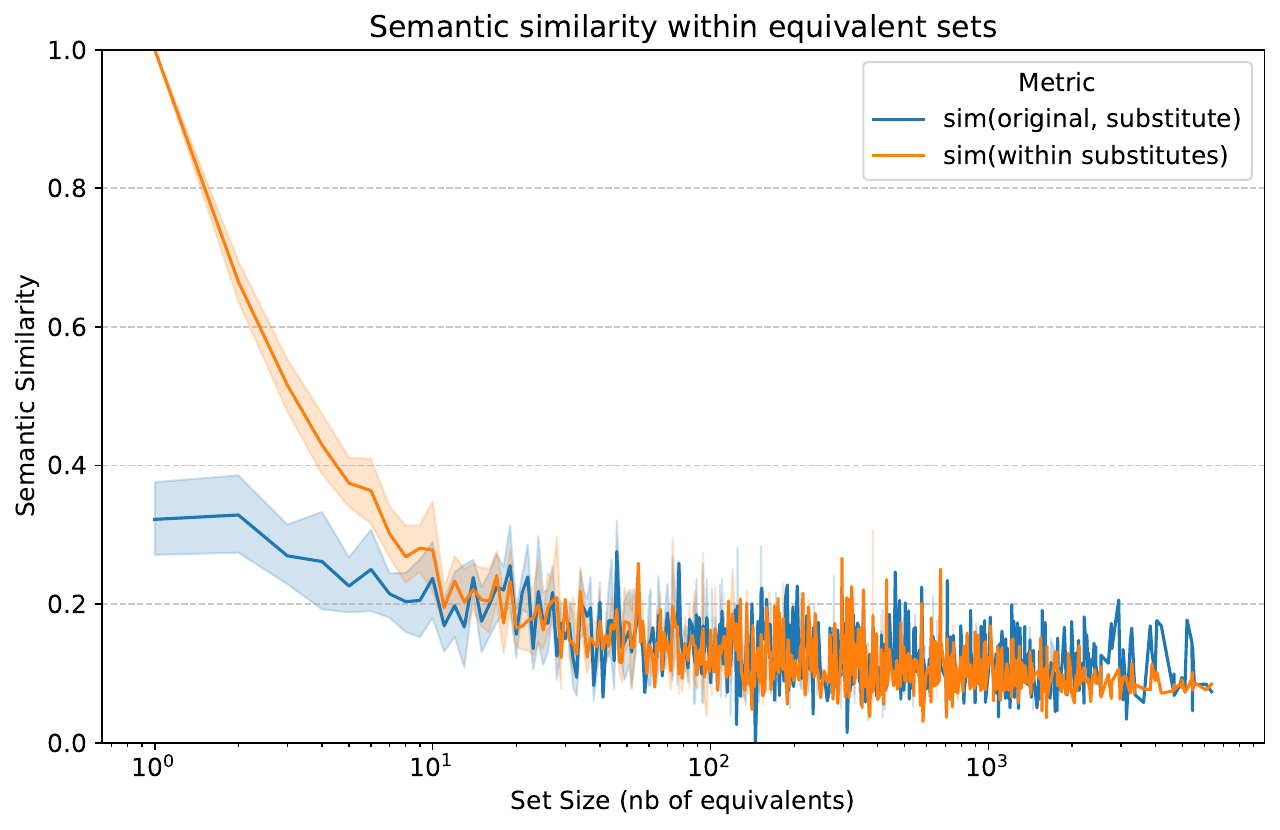}
    \end{subfigure}
    \caption*{(c) Semantic similarity between substitutes (\textit{left:} autoprompt; \textit{right:} original prompt).}

    \caption{\textbf{OLMo-7B}: Reproducing pruning and replacement experiments.}
    \label{fig:exp_olmo7b}
\end{figure*}

\begin{figure*}[t]
    \centering
    \begin{subfigure}{0.45\linewidth}
        \centering
        \includegraphics[width=\linewidth]{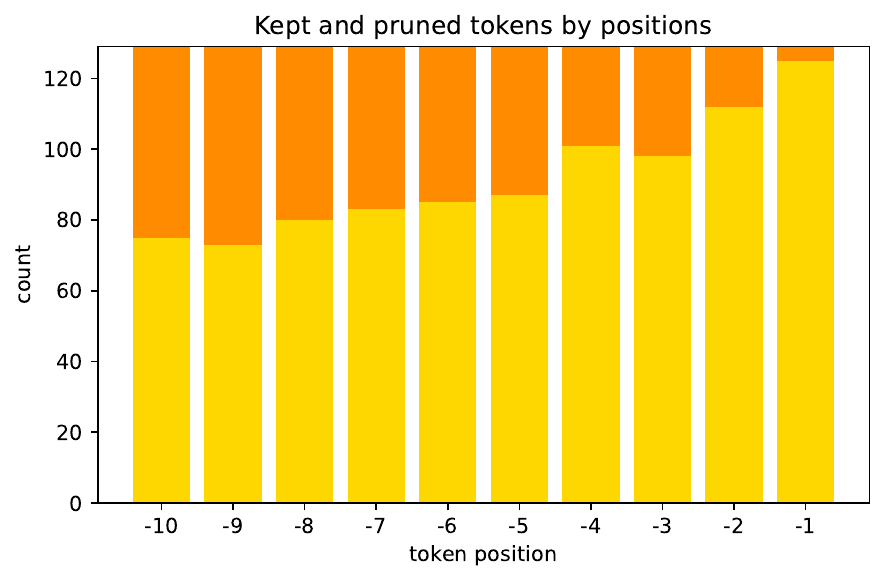}
    \end{subfigure}
    \hfill
    \begin{subfigure}{0.45\linewidth}
        \centering
        \includegraphics[width=\linewidth]{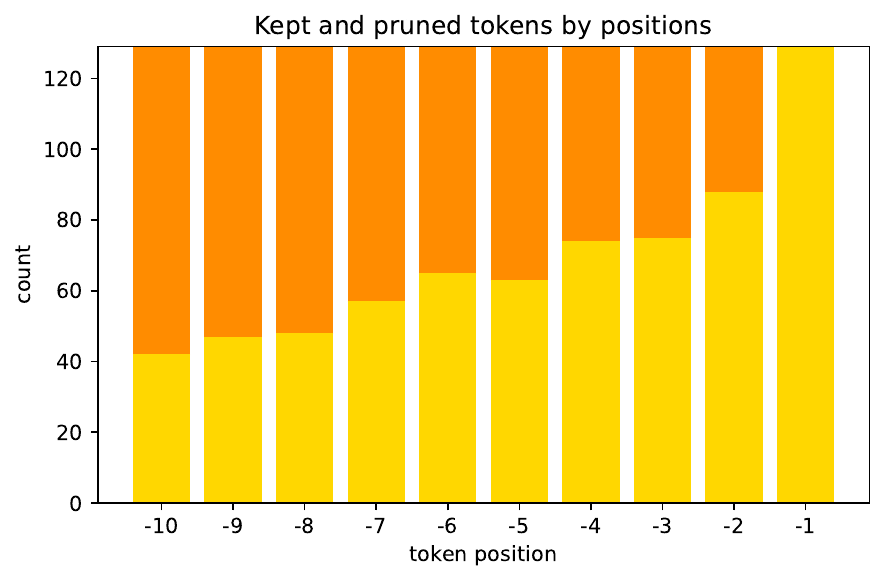}
    \end{subfigure}
    \caption*{(a) Count of pruned (orange) and kept (yellow) tokens (\textit{left:} autoprompt; \textit{right:} original prompt).}

    \begin{subfigure}{0.45\linewidth}
        \centering
        \includegraphics[width=\linewidth]{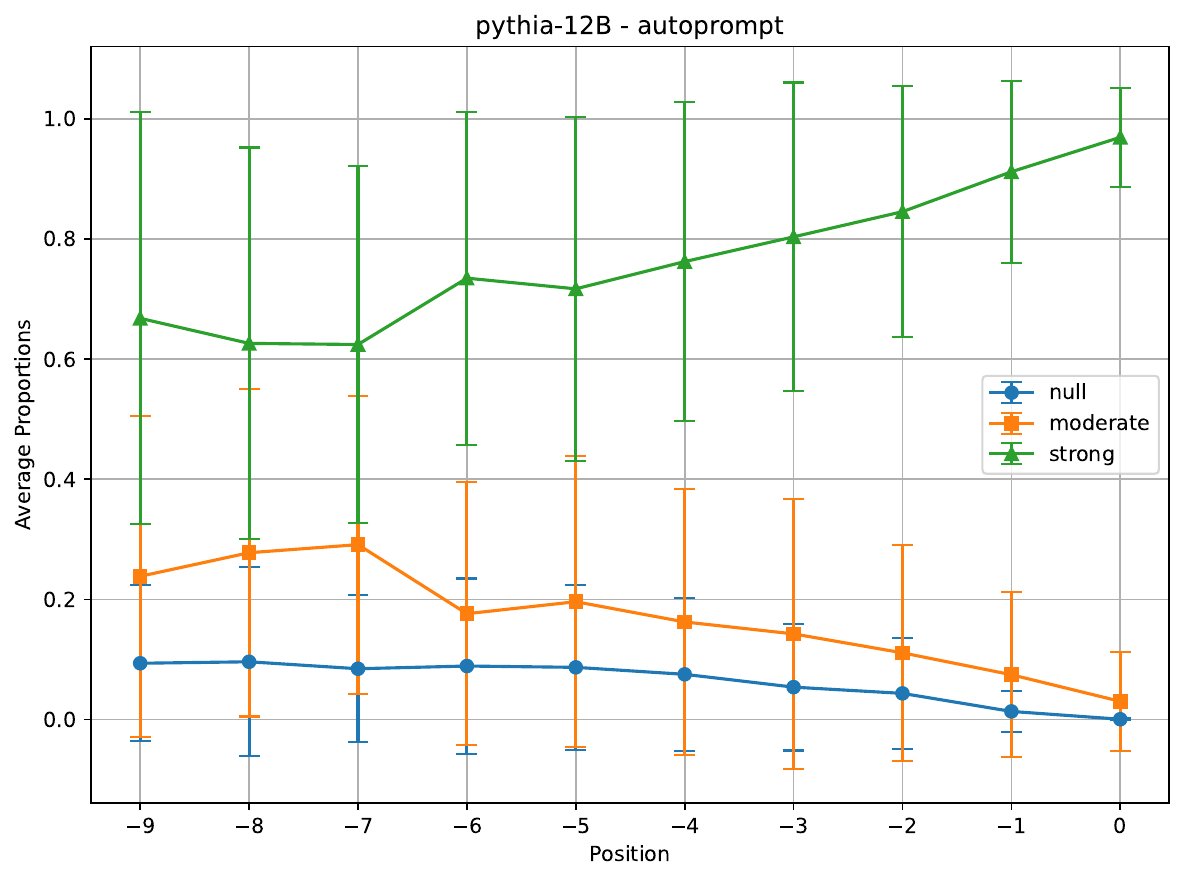}
    \end{subfigure}
    \hfill
    \begin{subfigure}{0.45\linewidth}
        \centering
        \includegraphics[width=\linewidth]{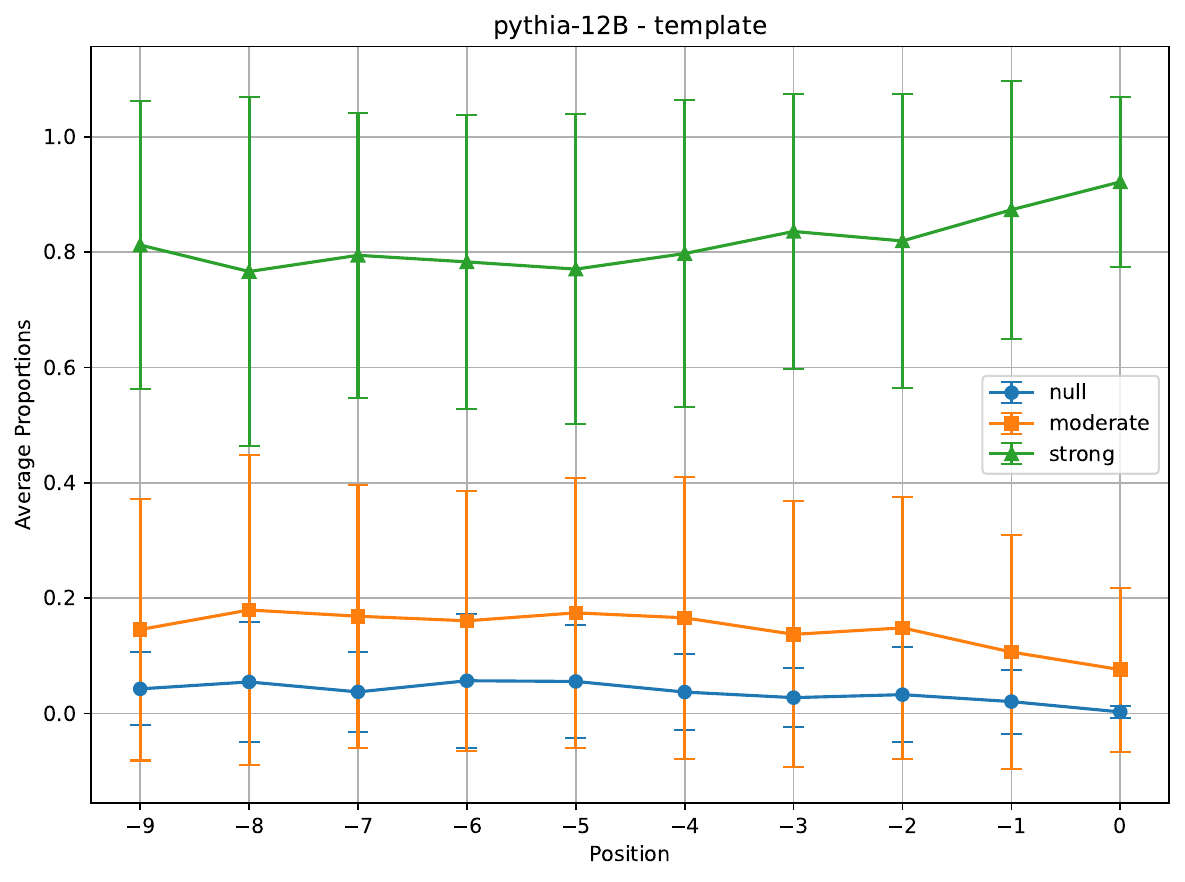}
    \end{subfigure}
    \caption*{(b) Average proportions of replacement effect types by position (\textit{left:} autoprompt; \textit{right:} original prompt).}

    \begin{subfigure}{0.45\linewidth}
        \centering
        \includegraphics[width=\linewidth]{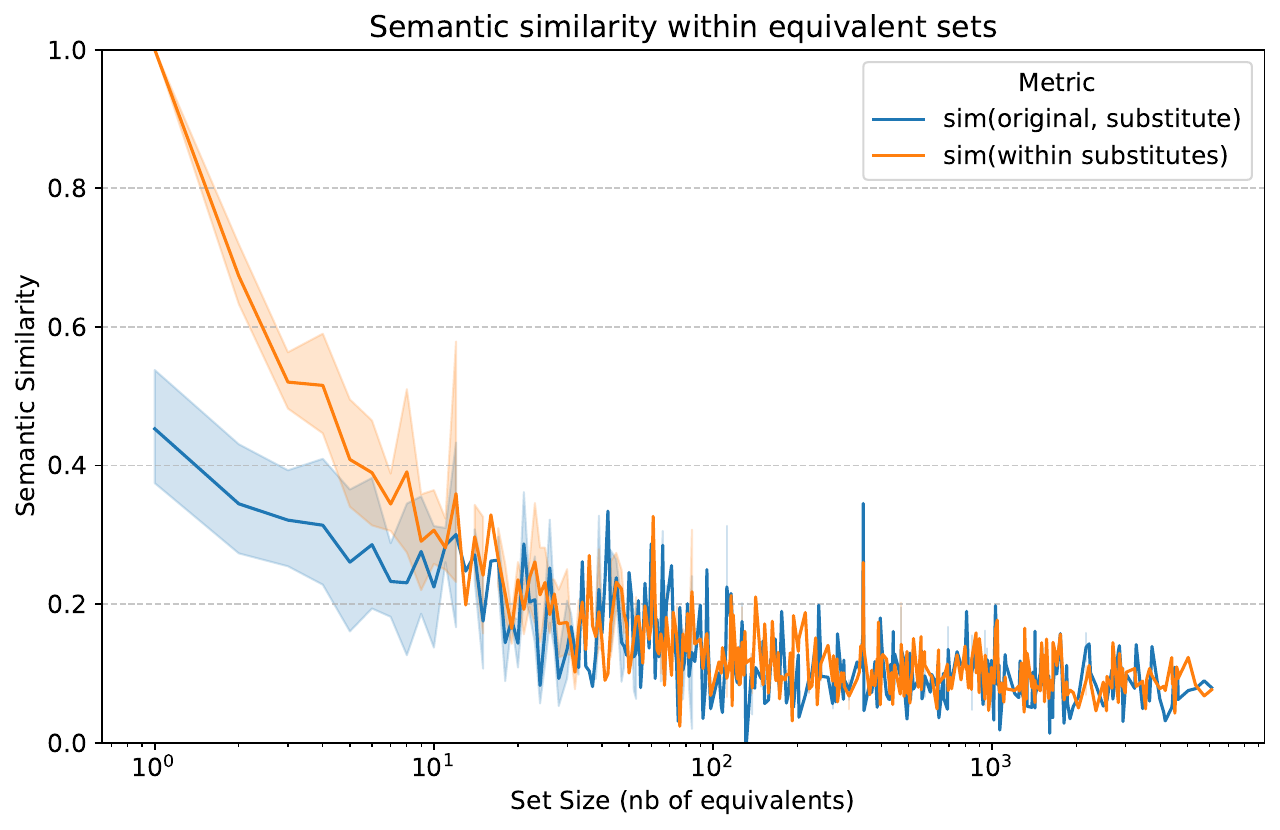}
    \end{subfigure}
    \hfill
    \begin{subfigure}{0.45\linewidth}
        \centering
        \includegraphics[width=\linewidth]{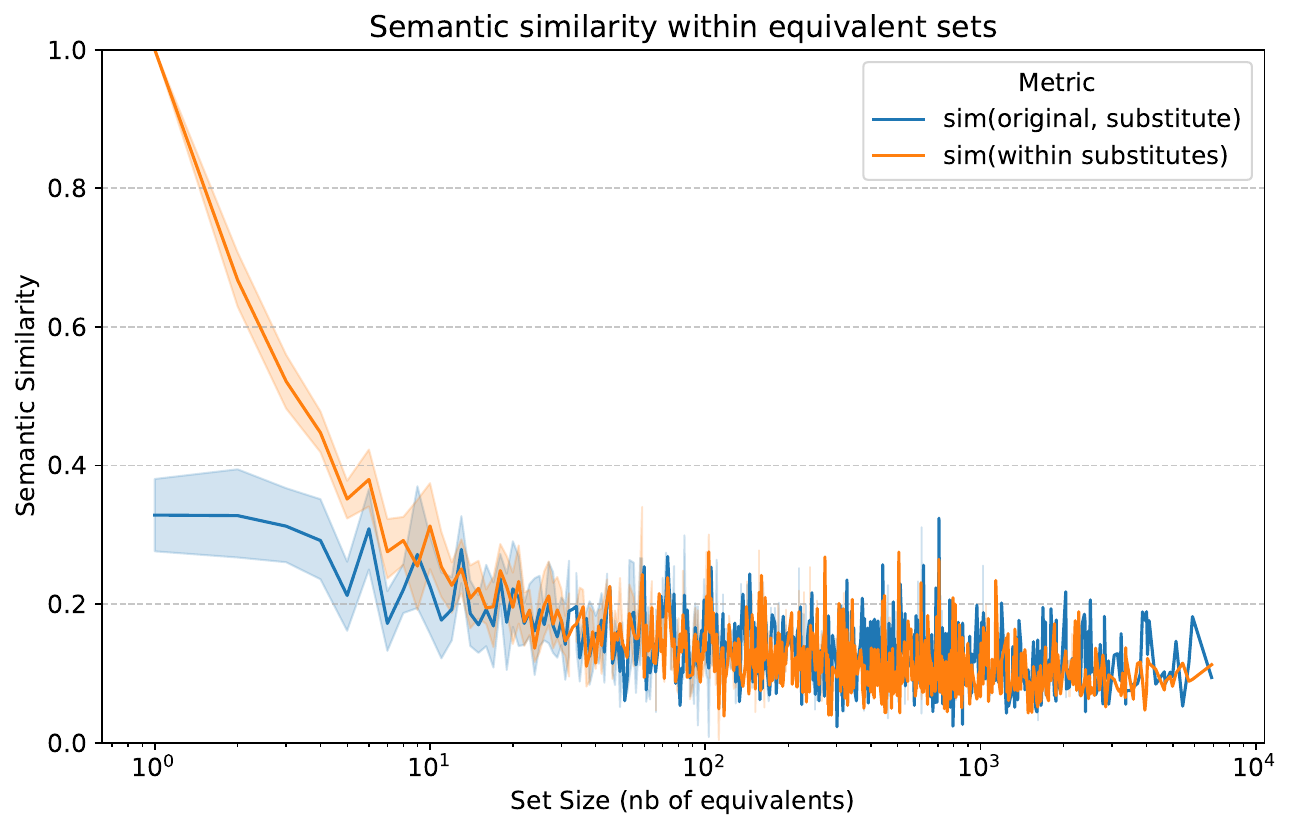}
    \end{subfigure}
    \caption*{(c) Semantic similarity between substitutes (\textit{left:} autoprompt; \textit{right:} original prompt).}

    \caption{\textbf{Pythia-12B}: Reproducing pruning and replacement experiments.}
    \label{fig:exp_pythia12b}
\end{figure*}

\begin{figure*}[t]
    \centering
    \begin{subfigure}{0.45\linewidth}
        \centering
        \includegraphics[width=\linewidth]{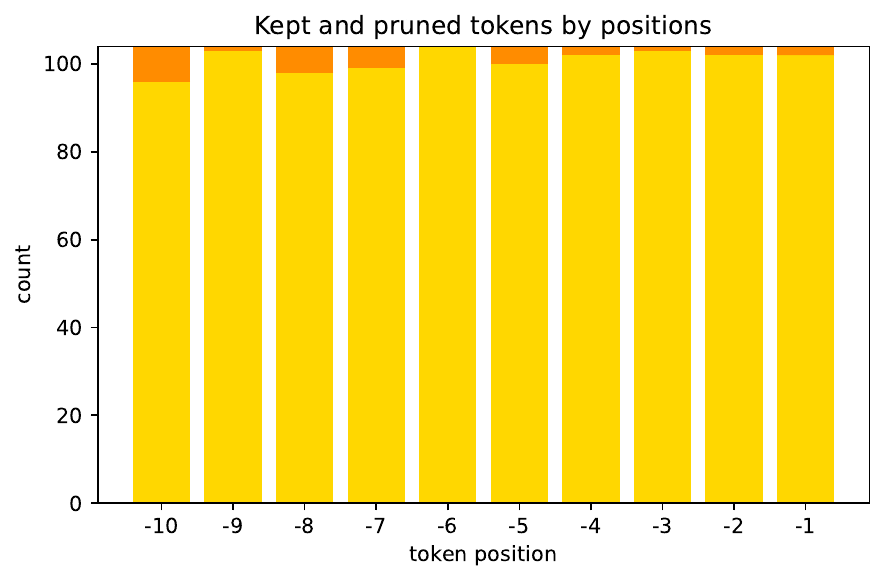}
    \end{subfigure}
    \hfill
    \begin{subfigure}{0.45\linewidth}
        \centering
        \includegraphics[width=\linewidth]{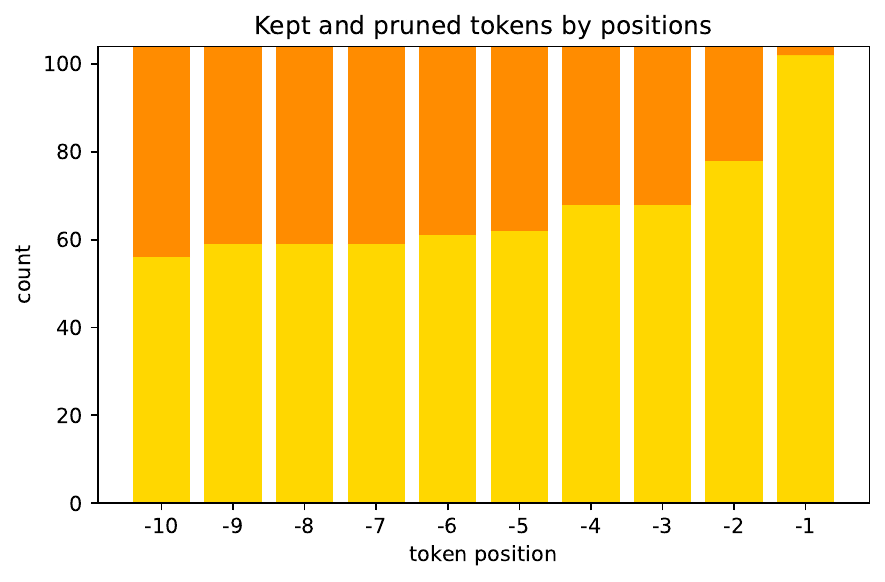}
    \end{subfigure}
    \caption*{(a) Count of pruned (orange) and kept (yellow) tokens (\textit{left:} autoprompt; \textit{right:} original prompt).}

    \begin{subfigure}{0.45\linewidth}
        \centering
        \includegraphics[width=\linewidth]{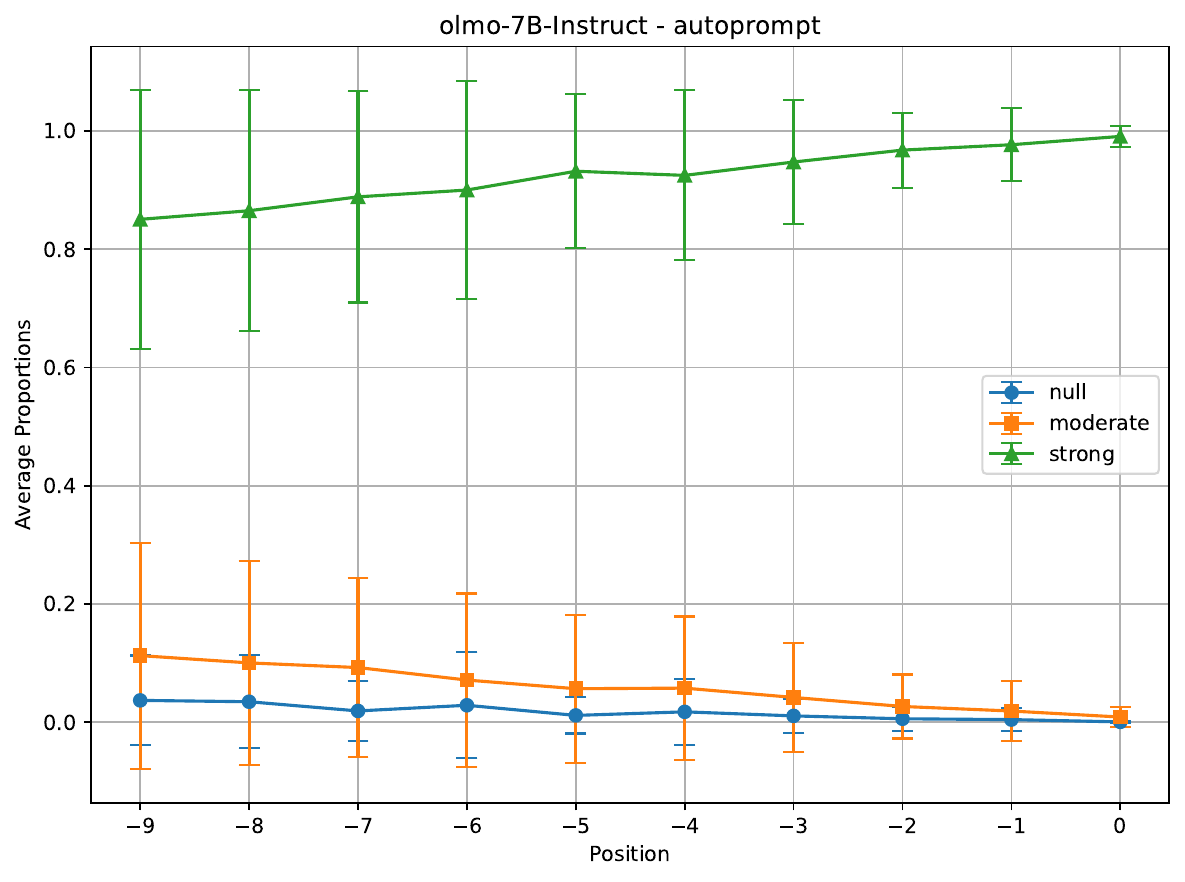}
    \end{subfigure}
    \hfill
    \begin{subfigure}{0.45\linewidth}
        \centering
        \includegraphics[width=\linewidth]{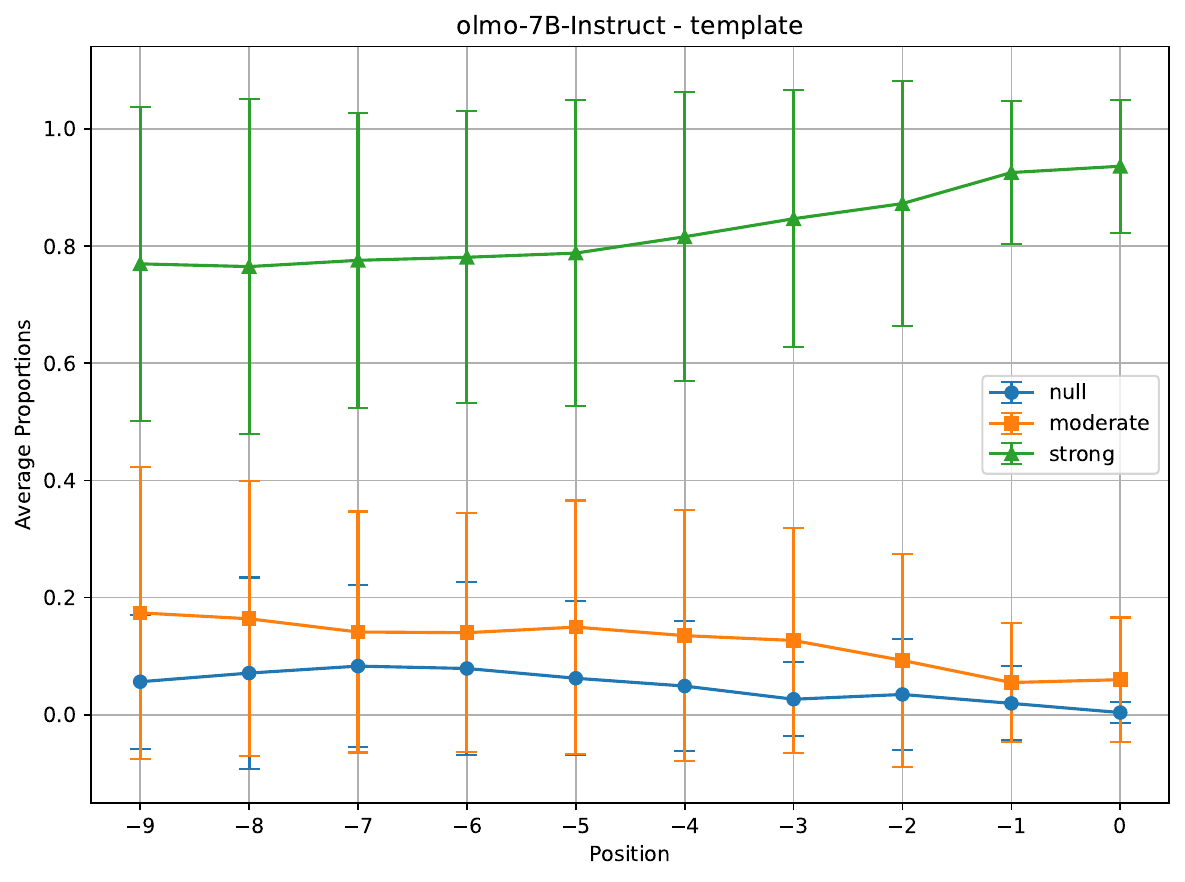}
    \end{subfigure}
    \caption*{(b) Average proportions of replacement effect types by position (\textit{left:} autoprompt; \textit{right:} original prompt).}

    \begin{subfigure}{0.45\linewidth}
        \centering
        \includegraphics[width=\linewidth]{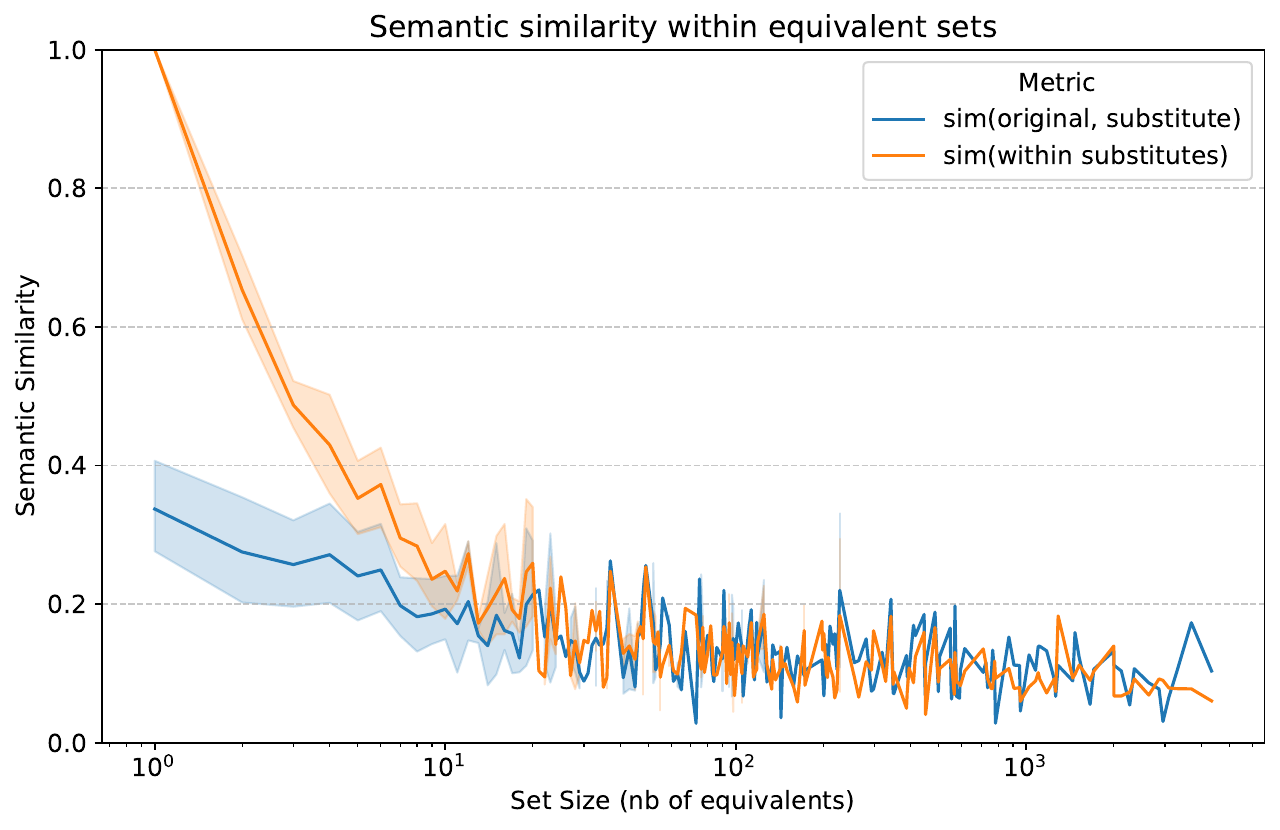}
    \end{subfigure}
    \hfill
    \begin{subfigure}{0.45\linewidth}
        \centering
        \includegraphics[width=\linewidth]{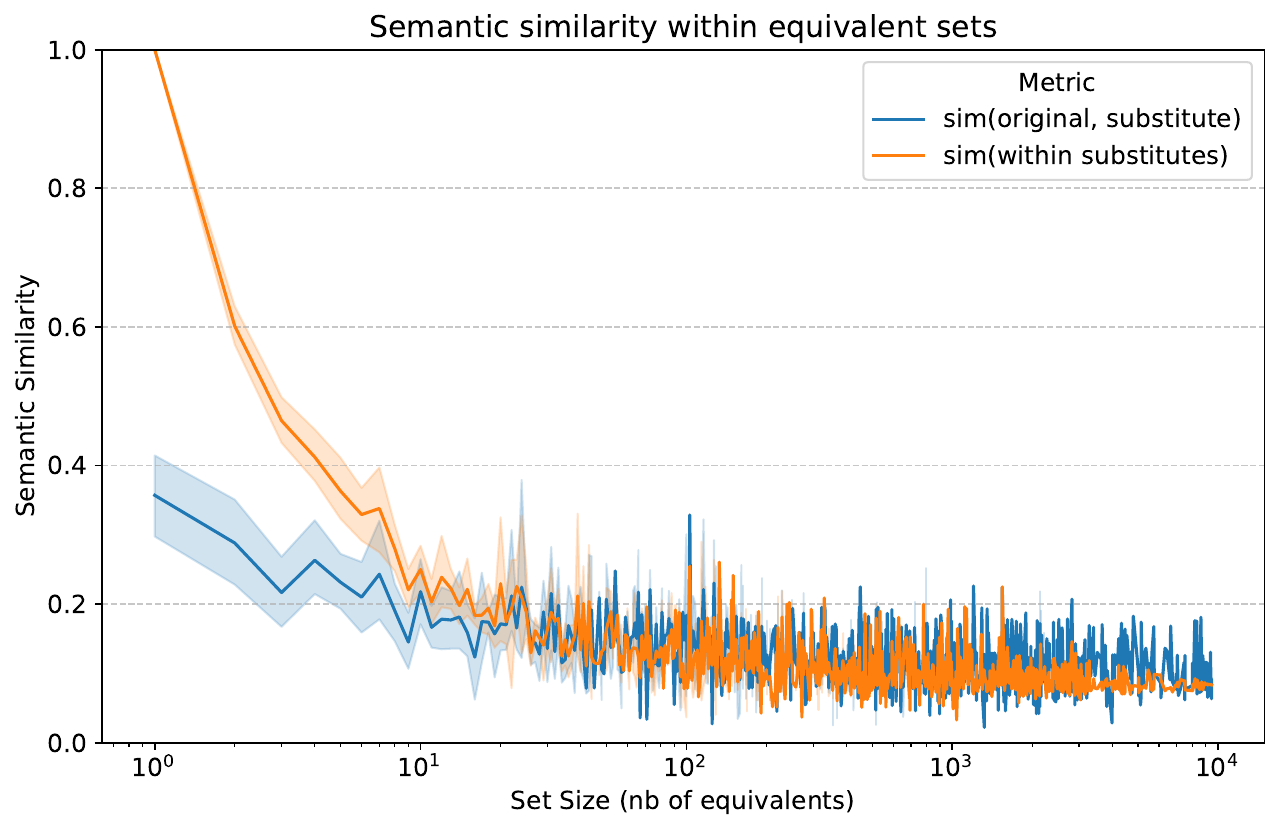}
    \end{subfigure}
    \caption*{(c) Semantic similarity between substitutes (\textit{left:} autoprompt; \textit{right:} original prompt).}

    \caption{\textbf{OLMo-7B-Instruct}: Reproducing pruning and replacement experiments.}
    \label{fig:exp_olmo7b_inst}
\end{figure*}

\paragraph{Pruning and shuffling}
The results of the pruning and shuffling experiments of natural prompts are presented in Table \ref{tab:natural_prompt_experiments}. For all models except OLMo-7B-Instruct, there is a difference in BLEU when shuffling all tokens vs. keeping last token fixed (paired t-test significant at p < 0.01). 
Token pruning by position is reported in Figures~\ref{fig:exp_pythia1.3b}~\ref{fig:exp_olmo1b}~\ref{fig:exp_olmo7b}~\ref{fig:exp_pythia12b}~\ref{fig:exp_olmo7b_inst} (a).

\paragraph{Replacement} Proportions of replacement effect type by position are reported in Figures~\ref{fig:exp_pythia1.3b}~\ref{fig:exp_olmo1b}~\ref{fig:exp_olmo7b}~\ref{fig:exp_pythia12b}~\ref{fig:exp_olmo7b_inst} (b).

\paragraph{Semantic consistency} Semantic consistency of the equivalent sets are reported in Figures~\ref{fig:exp_pythia1.3b}~\ref{fig:exp_olmo1b}~\ref{fig:exp_olmo7b}~\ref{fig:exp_pythia12b}~\ref{fig:exp_olmo7b_inst} (c).

\section{Token replacement examples}
\label{app:replacement-examples}

We show randomly picked examples of single-token autoprompt replacements that do not affect the continuation, have a moderate effect on it or have a strong effect on it in tables \ref{tab:replacement-examples-null}, \ref{tab:replacement-examples-moderate} and \ref{tab:replacement-examples-strong}, respectively. In addition, Table~\ref{tab:colored_sub_more} provides additional examples of replacement with null-effect, where tokens are colored based on the number of equivalent substitutes.

\arrayrulecolor{gray!50}
\begin{table*}[tb]
\begin{adjustbox}{width=\textwidth}
\begin{tabular}{l}
\textit{autoprompt:} crossesFootnote \sout{Several}/\textbf{Accordingly} panels accidentally have feather 517 chant collectively \\
\textit{continuation:} , as if they were a single bird. \\
\midrule
\textit{autoprompt:} <EOT><EOT> \sout{unbelievable}/\textbf{lands} deep intuitive \\
\textit{continuation:} understanding of the human condition. \\
\midrule
\textit{autoprompt:} throne were\sout{ivered}/\textbf{print}ceryassociated pursuing Somerset where \\
\textit{continuation:} he was arrested and imprisoned in the Tower of London. \\
\midrule
\textit{autoprompt:} \sout{tables}/\textbf{app} Its 590 chapel wide Rosadian \\
\textit{continuation:} marble floor is the largest in the world. \\
\midrule
\textit{autoprompt:} FLOAT depicts 1933\}\}?significant professional MMA \sout{classic}/\textbf{adapt} pickup after \\
\textit{continuation:} a long absence from the sport. \\
\end{tabular}
\end{adjustbox}
  \caption{Randomly selected \textit{null-effect} replacement examples. Replaced tokens and replacements are separated by ``/'' and in bold.}
  \label{tab:replacement-examples-null}
\end{table*}

\begin{table*}[tb]
\begin{tabular}{l}
\textit{autoprompt:} \sout{TLR}/\textbf{ract}<EOT> pledmanifest behalf breached adiabarfollow \\
\textit{original continuation:} \sout{the terms of the settlement agreement.} \\
\textit{modified continuation:} \textbf{ed the terms of the agreement.} \\
\textit{modified continuation BLEU:} 0.49 \\
\midrule
\textit{autoprompt:} stylish 1876 painting escaped creditors \sout{Anne}/\textbf{Im} decides \\
\textit{original continuation:} \sout{to sell it to pay off her debts.} \\
\textit{modified continuation:} \textbf{to sell it to a wealthy collector.} \\
\textit{modified continuation BLEU:} 0.32 \\
\midrule
\textit{autoprompt:} \sout{Round}/\textbf{22} transferring'\$, when splash energy \\
\textit{original continuation:}  \sout{is transferred to the surrounding water.} \\
\textit{modified continuation:} \textbf{is transferred to the surface of the liquid.} \\
\textit{modified continuation BLEU:} 0.32 \\
\midrule
\textit{autoprompt:} <EOT>\sout{<EOT>}/\textbf{Two}<EOT> Return onto Broadway where Gardens Airport coincide \\
\textit{original continuation:} \sout{with the Garden State Parkway.} \\
\textit{modified continuation:} \textbf{with the intersection of Broadway and the Garden State Parkway.} \\
\textit{modified continuation BLEU:} 0.45 \\
\midrule
\textit{autoprompt:} \sout{<EOT>}/\textbf{Where} Treaty<EOT>Unionduring \\
\textit{original continuation:} \sout{the American Revolutionary War.}\\
\textit{modified continuation:} \textbf{the Revolutionary War.}\\
\textit{modified continuation BLEU:} 0.33 \\
\end{tabular}
  \caption{Randomly selected \textit{moderate-effect} replacement examples (BLEU after replacement is of at least 0.2 but below 1). Replaced tokens and replacements are separated by ``/'' and in bold.}
  \label{tab:replacement-examples-moderate}
\end{table*}

\begin{table*}[tb]
\begin{tabular}{l}
\textit{autoprompt:} FIRST<EOT>protective talents\sout{Sarah}/\textbf{bias}<EOT> uses learning technical \\
\textit{original continuation:} \sout{skills to protect herself and others.} \\
\textit{modified continuation:} \textbf{skills to protect the user from harm.} \\
\textit{modified continuation BLEU:} 0.15 \\
\midrule
\textit{autoprompt:} fiveZero \sout{Parl}/\textbf{ins}terson homicide By 1867 Provincialswick warranted \\
\textit{original continuation:} \sout{the establishment of a police force.} \\
\textit{modified continuation:} \textbf{the appointment of a police magistrate.} \\
\textit{modified continuation BLEU:} 0.19 \\
\midrule
\textit{autoprompt:} stylish 1876 painting escaped \sout{creditors}/\textbf{()} Anne decides \\
\textit{original continuation:} \sout{to sell it to pay off her debts.} \\
\textit{modified continuation:} \textbf{to take a break from her work and go for a walk.} \\
\textit{modified continuation BLEU:} 0.02 \\
\midrule
\textit{autoprompt:} Pretty.“byterBlood "A realistic \sout{work}/\textbf{convicted} compatible \\
\textit{original continuation:} \sout{with the other works of the author} \\
\textit{modified continuation:} \textbf{with the "Atheist" and "Atheist" "Atheist" an[...]} \\
\textit{modified continuation BLEU:} 0.02 \\
\midrule
\textit{autoprompt:}  ------------------------------------------------------terrorismworker \sout{is}/\textbf{exit} killed \\
\textit{original continuation:} \sout{in a suicide bombing in Iraq.} \\
\textit{modified continuation:} \textbf{in Iraq} \\
\textit{modified continuation BLEU:} 0.03 \\
\end{tabular}
  \caption{Randomly selected \textit{strong-effect} replacement examples (BLEU after replacement is below 0.2). Replaced  tokens and replacements are separated by ``/'' and in bold. Hard-to-render characters replaced by ``?''.}
  \label{tab:replacement-examples-strong}
\end{table*}

\arrayrulecolor{gray!50}
\begin{table*}[th]
    \centering
    \small  %
    \renewcommand{\arraystretch}{0.91}  %
    \begin{tabularx}{\linewidth}{p{7cm}X}
        \specialrule{1pt}{0pt}{0pt}
                \rowcolor{gray!20} \textbf{Autoprompts} & \textbf{Generated Continuation}   \\
        \specialrule{1pt}{0pt}{2pt}
        \secondbox{ipzig}  
        \secondbox{suspended}  
        \secondbox{Emma}  
        \firstbox{Leigh} \firstbox{'s} \firstbox{956} \firstbox{time}  
        \secondbox{commitments}  
        \secondbox{and}  
        \secondbox{as}  
        & \dots a result, she was able to devote herself to her writing. \\
        \noalign{\vskip 0.5ex}\hdashline\noalign{\vskip 0.5ex}
        \multicolumn{2}{l}{\textbf{ipzig}: \textit{' century', ' society', ' secret', ' glad', ' revolution', ' intellectual', ' consciousness', \dots}}\\
        \multicolumn{2}{l}{\textbf{Emma}: \textit{' women', ' woman', ' tau', ' wife', ' sex', ' da', ' ura', ' snow', \dots}}\\
        \multicolumn{2}{l}{\textbf{commitments}: \textit{' attention', ' traditional', ' burden', ' politics', ' busy', \dots}}\\
        \multicolumn{2}{l}{\textbf{and}: \textit{' how', ' and', ' et', ' that', ' ad', ' und', \dots}}\\
        \multicolumn{2}{l}{\textbf{as}: \textit{' as'}}\\
        \multicolumn{2}{l}{\textbf{suspended}: \textit{0 substitutes}} \\ \\
        \midrule
        \firstbox{COMMAND} \firstbox{uminous}  
        \secondbox{photometry}  
        \secondbox{has}  
        \firstbox{?} \firstbox{continually}  
        \secondbox{undermined}  
        & \dots the idea that the Sun is a star. \\
        \noalign{\vskip 0.5ex}\hdashline\noalign{\vskip 0.5ex}
        \multicolumn{2}{l}{\textbf{has}: \textit{' helps', ' raise', ' Williams'}}\\
        \multicolumn{2}{l}{\textbf{undermined}: \textit{' rejected'}}\\
        \multicolumn{2}{l}{\textbf{photometry}: \textit{0 substitutes.}} \\ 
        \specialrule{1pt}{0pt}{0pt}
                \rowcolor{gray!20} \textbf{Original Prompt} & \textbf{Generated Continuation} \\
        \specialrule{1pt}{0pt}{2pt}
         \firstbox{South} \firstbox{Australian} \firstbox{with} \firstbox{Wales} 
         \secondbox{requested} 
         \secondbox{that} 
         \secondbox{Major} \secondbox{General} 
         \firstbox{William} \firstbox{ois}
         & be appointed to command the expedition.\\
        \hdashline
         \multicolumn{2}{l}{\textbf{requested}: \textit{' asked'}}\\
         \multicolumn{2}{l}{\textbf{Major}: \textit{' sub',' major',' officer',' Sub'}}\\
         \multicolumn{2}{l}{\textbf{General}: \textit{' general',' officer',' Gen',' Captain'}}\\
         \multicolumn{2}{l}{\textbf{that}: \textit{0 substitutes.}} \\
         \midrule
         \firstbox{erected} 
        \firstbox{in} \firstbox{1896} \firstbox{listed} \firstbox{of} 
        \secondbox{the} 
        \firstbox{sand} 
        \secondbox{stone} 
        \firstbox{market} 
        \secondbox{building}
         &, which was built in the late 19th century.\\
         \midrule
         \multicolumn{2}{l}{\textbf{the}: \textit{' an',' its',' some',' great',' old',' further',' additional',' especially', \dots}}\\
         \multicolumn{2}{l}{\textbf{stone}: \textit{'aligned','ized','á','ette','ished',' stone','split','white','ulate', \dots}}\\
         \multicolumn{2}{l}{\textbf{building}: \textit{' court',' structure',' complex',' House',' unit',' track',' society',' mechanism', \dots}}\\
        \bottomrule
    \end{tabularx}
    \caption{Additional Examples of replacement for autoprompts and original prompts (10 last tokens). Color represents the number of substitutes: \firstbox{$>50$ substitutes} and \secondbox{$<50$ substitutes}. 
    When there are less than 50 substitutes, a random subset is displayed. Difficult-to-render characters are replaced by '?'.}
    \label{tab:colored_sub_more}
\end{table*}
\arrayrulecolor{black}

\paragraph{Large substitution sets} Table~\ref{tab:dark_matter_full} presents a  larger list of tokens that are the most frequently found in large equivalent substitution sets. It mostly consists of subwords (sometimes corresponding to interpretable morphemes), digits and (fragments of) named entities, that do not carry strong semantic information. We conjecture that digits appear in cases where the exact number is not crucial for generating the continuation, but the notion of quantity or date is important (e.g., in that case $5$, $10$ or $18$ are equivalents). A similar interpretation holds for named entities. For instance, `Moore', `Brown' or `Smith' are common proper nouns that could refer to a typical North-American entity in an exchangeable way.

\begin{table*}[h]
    \centering
    \renewcommand{\arraystretch}{1.2}
    \setlength{\tabcolsep}{3pt}
    \footnotesize
    \begin{adjustbox}{width=1\textwidth}
    \begin{tabular}{|ccccccccccccccc|}
    \hline
    \firstbox{105} & \secondbox{<EOT>} & \secondbox{()} & \firstbox{113} & \fourthbox{ini} & \thirdbox{Mar} & \fourthbox{ran} & \firstbox{25} & \fourthbox{be} & \thirdbox{Green} & \firstbox{101} & \firstbox{56} & \secondbox{¸} & \fourthbox{mat} & \fourthbox{ato} \\
    \fourthbox{ator} & \thirdbox{Rock} & \firstbox{35} & \firstbox{115} & \thirdbox{Moore} & \firstbox{114} & \fourthbox{SS} & \fourthbox{lim} & \fourthbox{eter} & \fourthbox{ester} & \fourthbox{abe} & \fourthbox{uns} & \thirdbox{Head} & \thirdbox{Char} & \secondbox{?} \\
    \secondbox{?} & \secondbox{?} & \fourthbox{ac} & \fourthbox{aa} & \fourthbox{aur} & \secondbox{?} & \fourthbox{ong} & \fourthbox{ost} & \fourthbox{den} & \fourthbox{FF} & \firstbox{54} & \fourthbox{a} & \fourthbox{ab} & \fourthbox{ud} & \fourthbox{sl} \\
    \fourthbox{rim} & \fourthbox{st} & \secondbox{?} & \firstbox{022} & \firstbox{1} & \firstbox{37} & \firstbox{91} & \fourthbox{INT} & \fourthbox{op} & \fourthbox{ore} & \fourthbox{ef} & \fourthbox{rel} & \fourthbox{cha} & \fourthbox{ast} & \fourthbox{ives} \\
    \firstbox{2000} & \fourthbox{ash} & \fourthbox{arch} & \fourthbox{ath} & \fourthbox{eps} & \fourthbox{ced} & \thirdbox{Har} & \fourthbox{fall} & \fourthbox{met} & \fourthbox{ales} & \firstbox{140} & \thirdbox{Ros} & \firstbox{5} & \firstbox{48} & \firstbox{21} \\
    \firstbox{33} & \firstbox{101} & \firstbox{38} & \firstbox{75} & \firstbox{600} & \secondbox{?} & \fourthbox{**} & \fourthbox{oth} & \fourthbox{aff} & \fourthbox{ay} & \fourthbox{br} & \fourthbox{LE} & \secondbox{?} & \fourthbox{anch} & \fourthbox{der} \\
    \fourthbox{cy} & \fourthbox{ris} & \fourthbox{fer} & \fourthbox{inner} & \fourthbox{ots} & \fourthbox{arn} & \thirdbox{Sil} & \thirdbox{Red} & \fourthbox{ros} & \secondbox{,} & \fourthbox{cent} & \thirdbox{Brown} & \fourthbox{ister} & \fourthbox{dec} & \firstbox{50} \\
    \firstbox{44} & \firstbox{40} & \firstbox{28} & \firstbox{500} & \secondbox{?} & \firstbox{41} & \firstbox{46} & \firstbox{85} & \firstbox{112} & \firstbox{93} & \firstbox{45} & \firstbox{57} & \fourthbox{n} & \fourthbox{y} & \fourthbox{em} \\
    \fourthbox{ard} & \fourthbox{na} & \fourthbox{eng} & \fourthbox{ater} & \fourthbox{ree} & \fourthbox{EE} & \fourthbox{urg} & \secondbox{?} & \secondbox{?} & \fourthbox{ord} & \fourthbox{oto} & \firstbox{1} & \secondbox{-} & \fourthbox{ign} & \fourthbox{sen} \\
    \fourthbox{ans} & \fourthbox{ach} & \fourthbox{dis} & \fourthbox{ush} & \fourthbox{Br} & \thirdbox{Cam} & \fourthbox{pol} & \fourthbox{one} & \secondbox{?} & \fourthbox{ner} & \firstbox{18} & \firstbox{42} & \firstbox{48} & \firstbox{250} & \firstbox{96} \\
    \thirdbox{Bell} & \firstbox{94} & \firstbox{117} & \firstbox{19} & \fourthbox{ial} & \fourthbox{pp} & \fourthbox{ba} & \fourthbox{star} & \secondbox{?} & \fourthbox{mar} & \fourthbox{har} & \fourthbox{hor} & \thirdbox{Tor} & \fourthbox{lan} & \fourthbox{vi} \\
    \fourthbox{Sl} & \firstbox{1986} & \fourthbox{arg} & \fourthbox{ender} & \thirdbox{Sand} & \thirdbox{First} & \thirdbox{Po} & \thirdbox{Che} & \thirdbox{Jones} & \firstbox{100} & \fourthbox{per} & \fourthbox{ative} & \fourthbox{ways} & \fourthbox{hal} & \fourthbox{ben} \\
    \fourthbox{bon} & \thirdbox{Ross} & \firstbox{6} & \secondbox{?} & \firstbox{30} & \firstbox{46} & \firstbox{31} & \firstbox{65} & \firstbox{400} & \firstbox{1000} & \fourthbox{LE} & \firstbox{33} & \fourthbox{am} & \thirdbox{Bar} & \fourthbox{ace} \\
    \fourthbox{amb} & \thirdbox{Ab} & \fourthbox{ull} & \thirdbox{Reg} & \fourthbox{ena} & \fourthbox{orm} & \fourthbox{stra} & \fourthbox{s} & \fourthbox{co} & \fourthbox{ards} & \fourthbox{win} & \fourthbox{iver} & \thirdbox{Ha} & \secondbox{?} & \fourthbox{pin} \\
    \fourthbox{init} & \fourthbox{dist} & \fourthbox{pres} & \thirdbox{Ray} & \firstbox{47} & \thirdbox{Smith} & \firstbox{120} & \firstbox{128} & \fourthbox{number} & \fourthbox{án} & \thirdbox{Kar} & \firstbox{4} & \firstbox{70} & \firstbox{26} & \firstbox{27} \\
    \firstbox{98} & \firstbox{200} & \firstbox{80} & \firstbox{200} & \firstbox{2000} & \firstbox{45} & \firstbox{500} & \firstbox{49} & \firstbox{110} & \firstbox{63} & \firstbox{13} & \fourthbox{el} & \fourthbox{ale} & \fourthbox{za} & \fourthbox{ute} \\
    \hline
    \end{tabular}
    \end{adjustbox}
    \caption{List of the $240$ tokens most frequently found in large equivalent sets (observed in about $10\%$ of the sets).  Difficult-to-render characters are replaced by ’?’. Tokens are marked as follows: \firstbox{digit}, \secondbox{artifact and punctuation}, \thirdbox{named entity}, \fourthbox{subword and morphological unit.}}
    \label{tab:dark_matter_full}
\end{table*}

\begin{figure*}[tb]
    \centering
    \begin{subfigure}{0.40\linewidth}
        \centering
        \includegraphics[width=\linewidth]{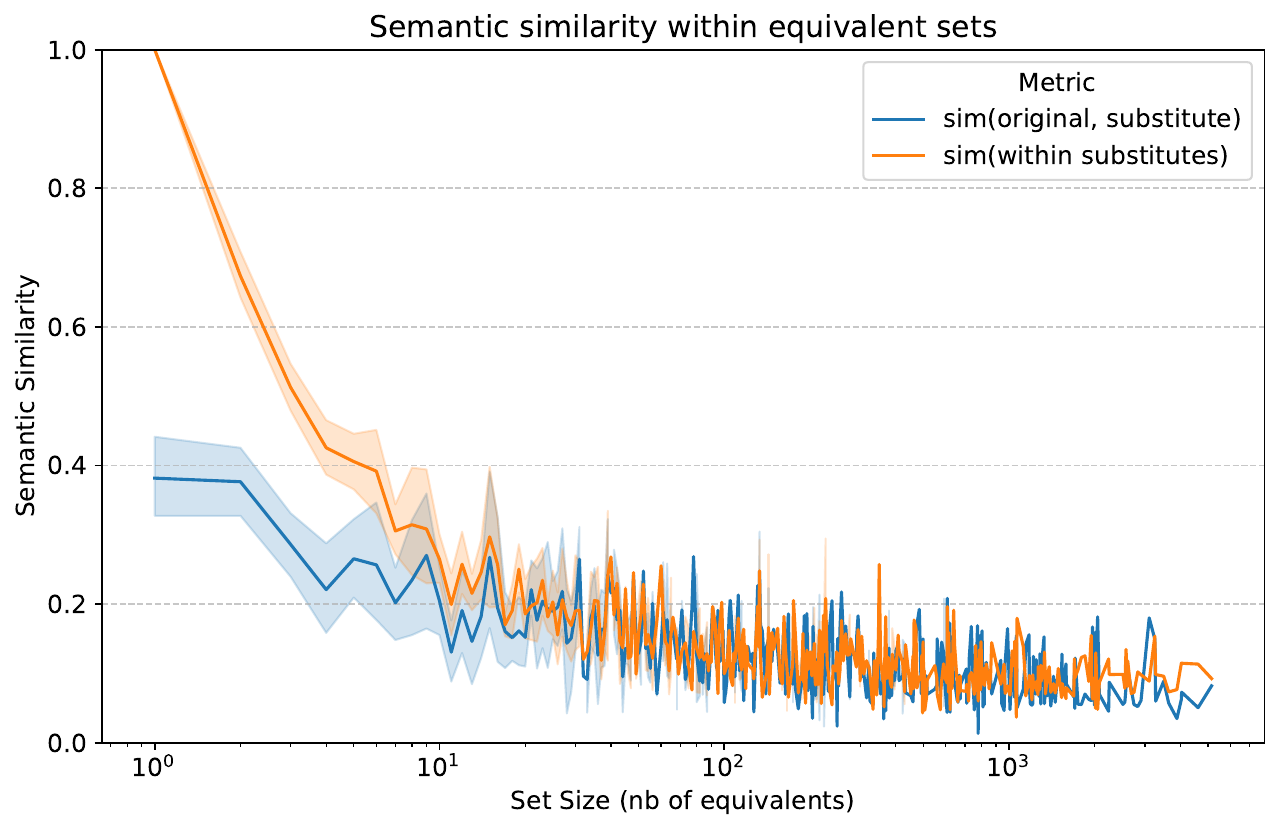}
        \captionsetup{aboveskip=0pt} %
        \caption{Autoprompts}
        \label{fig:sem_sub}
    \end{subfigure}
    \hfill
    \begin{subfigure}{0.40\linewidth}
        \centering
        \includegraphics[width=\linewidth]{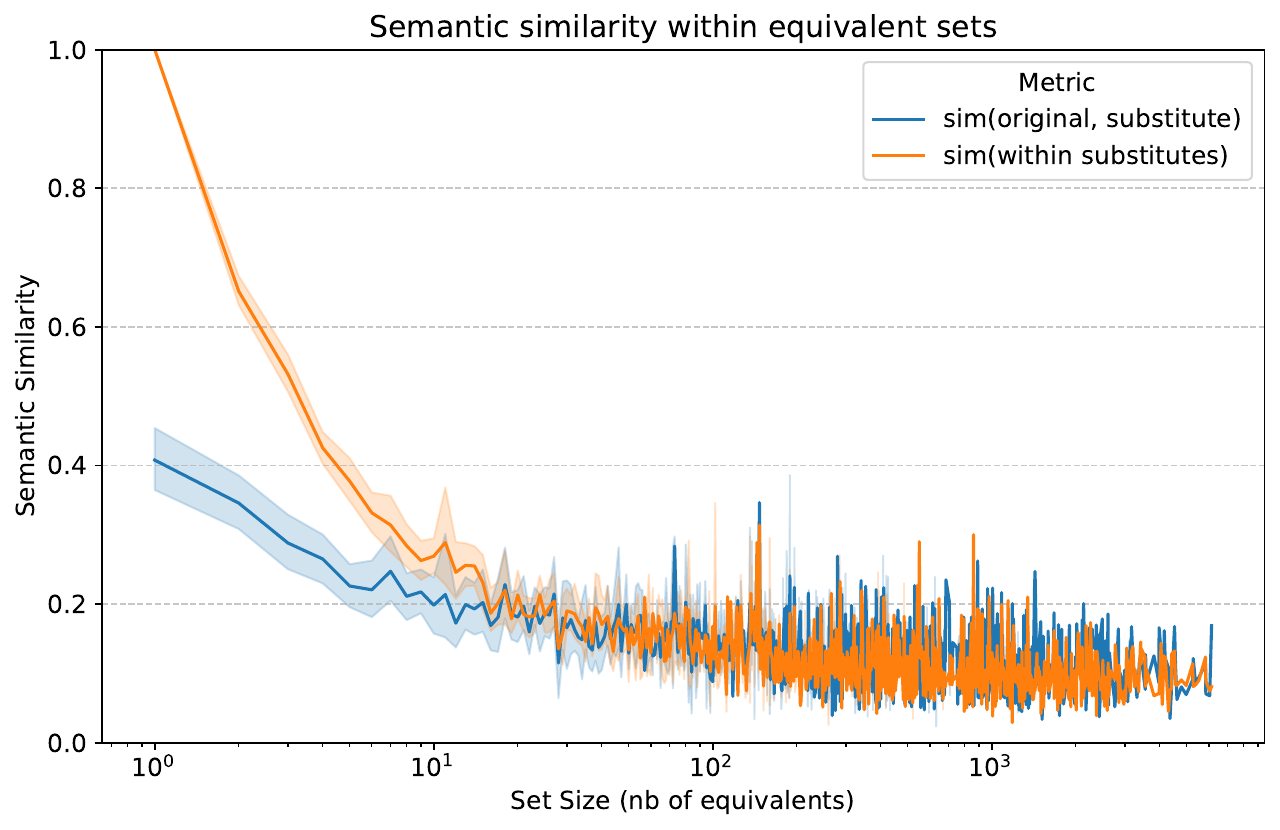}
        \captionsetup{aboveskip=0pt} %
        \caption{Original Prompts}
        \label{fig:sem_sub_template}
    \end{subfigure}
    \caption{Semantic similarity between substitutes. When a token admits fewer substitutes, they generally lie in a smaller semantic space than when the substitute count is large. The rightmost data points correspond to the most flexible tokens and tend to be semantically inconsistent. The leftmost data points are tokens admitting 0 or few substitutes, which are semantically closer.}%
    \label{fig:sem_sum_main}
\end{figure*}

\paragraph{Semantic consistency of the equivalent sets}
\label{app:sem}
Table~\ref{fig:sem_sum_main} provides the semantic similarity between substitutes.

\section{Human annotation of token importance}
\label{app:annotation}
The four authors of this paper manually annotated the autoprompt dataset (for the Pythia 6.9B model), assigning a binary label to each token to indicate whether it is intuitively important for generating the continuation. The annotation process was conducted \textit{a posteriori}, with annotators having access to the full autoprompt and its corresponding generated continuation. To minimize order effects, tokens were randomized before annotation. As an illustration, Table~\ref{tab:annotation} provides some annotated examples.

In total, $1971$ tokens  have been labeled by at least one annotator. We assess inter-annotator agreement on a subset of $50$ tokens: the four annotators agreed on $63.3\%$ of the examples, and at least 3 annotators agreed on $89.8\%$ of the examples.

Table~\ref{tab:human_prediction} provides the full comparison between importance annotation and equivalent set size.

\begin{table*}[]
    \centering
    \begin{adjustbox}{width=\textwidth}
    \begin{tabular}{llcc}
            \specialrule{1pt}{0pt}{0pt}
        \rowcolor{gray!20} \textbf{Autoprompt} & \textbf{Continuation} & \textbf{Token to annotate} & \textbf{Annotation} \\
    \specialrule{1pt}{0pt}{2pt}
        <EOT><EOT><EOT><EOT> \_\_\_\_\_<EOT> European devastated 1941 prompted & \dots the United States to enter the war. & <EOT> &	0\\
        clut British'); from 40 UTC anyone emitted one speaker & \dots in the UK would be heard in the UK. & speaker &	1 \\
        Croat trainersers?? Greece Catholics Deltaclaim independently & \dots from the Greek Orthodox Church. & Greece& 1\\
        Shan Marxist Augustine Fran??ois State TERis Railroad serves ? & \dots the city of Terre Haute, Indiana. & Railroad & 1\\
        defenseLittle<EOT>Content<EOT> forests constitute reliable leaving habitats & \dots for many species of birds and mammals. & leaving & 1\\
        earliest 2??–¢ flares when Jacksonville dominancethe metropolitan & \dots area was still in its infancy. & dominance & 0\\
        ‡?Å Berry paternalEffects cephal???. Johann Born evenublished & \dots a book on the subject in 1775. & Johann & 1\\
        1953 that Dix Teachers 1989 Archbishopreceived a Hamilton Legacy & \dots Award for his contributions to education. & that & 0\\
        \bottomrule
    \end{tabular}
    \end{adjustbox}
    \caption{We provide 8 examples of human annotations of token's importance. Using the autoprompt and the continuation, the annotator had to label the ``token to annotate'' as \textit{important} (labeled with 1) or \textit{not important} (labeled with 0). Difficult-to-render characters are replaced by ’?’}
    \label{tab:annotation}
\end{table*}
\arrayrulecolor{black}

\begin{table*}[tb]
    \centering
    \setlength{\tabcolsep}{6pt} %
    \sisetup{table-format=3.1} %
    \begin{tabular}{lS[table-format=3.1]S[table-format=3.1]S[table-format=3.1]S[table-format=3.1]S[table-format=3.1]S[table-format=3.1]}
        \toprule
        & \multicolumn{2}{c}{All} & \multicolumn{2}{c}{Lang-like} & \multicolumn{2}{c}{Non-Linguistic} \\
        \cmidrule(lr){2-3} \cmidrule(lr){4-5} \cmidrule(lr){6-7}
        & \textbf{Imp.} & \textbf{Not-Imp.} & \textbf{Imp.} & \textbf{Not-Imp.} & \textbf{Imp.} & \textbf{Not-Imp.} \\
        \midrule
        Average & 163.4 & 372.2 & 150.8 & 298.7 & 268.6 & 538.4 \\
        Median  & 2.0   & 35.0  & 2.0   & 23.0  & 13.5  & 83.0  \\
        \bottomrule
    \end{tabular}
    \caption{Comparison of the equivalent set size (average and median) between tokens classified as Important (Imp.) or Not-Important (Not-Imp.) by human experts, across different token categories.}
    \label{tab:human_prediction}
\end{table*}

\section{Computing resources}
All experiments were run on a cluster composed of 11 nodes with 5 NVIDIA A30 GPUs each. The autoprompt search for Pythia-1.4B took approximately 600 GPU hours. Pruning, replacement and shuffling experiments for Pythia-1.4B took 1500 GPU hours overall. Compute demand for the other models was comparable, although we had to settle for smaller autoprompt sets.

\section{Assets}
Besides standard tools such as Python and libraries such as NumPy and SciPy, we used the following tools and datasets, in accordance with their respective terms and licenses. 
\begin{itemize}
\item Dolma \url{https://huggingface.co/datasets/allenai/dolma}; license: ODC-By
\item NLTK \url{https://www.nltk.org}; license: apache-2.0
\item OLMo \url{https://huggingface.co/allenai/OLMo-7B}; license: apache-2.0
\item The Pile \url{https://pile.eleuther.ai/}; license: MIT
\item PyTorch \url{https://pytorch.org}; license: bsd
\item Pythia \url{https://huggingface.co/EleutherAI/pythia-1.4b-deduped}; license:
apache-2.0 
\item Huggingface Transformers \url{https://github.com/huggingface/transformers}; license:
apache-2.0
\item Wikitext \url{https://huggingface.co/datasets/wikitext}; license: Creative Commons
Attribution Share Alike 3.0

\end{itemize}

\end{document}